%% file: main.tex
\documentclass[final]{siamltex}
\usepackage{algorithm,algpseudocode}
\usepackage{latexsym, graphicx, epsfig, amsmath, amsfonts,amssymb,bm}
\usepackage{caption, subcaption}
\usepackage{color,soul}
\usepackage{epstopdf}
\usepackage[font=small,skip=0pt]{caption}
\usepackage{booktabs}
\usepackage{hyphenat}
\usepackage{multirow}
\usepackage{array}
\usepackage{color}
\usepackage{url}
\usepackage{comment}
\usepackage{diagbox}
\usepackage{graphicx}
\usepackage[label font=bf,labelformat=simple]{subfig}
\usepackage{caption}

\allowdisplaybreaks

\def\be{\begin{equation}}
\def\ee{\end{equation}}

\def\0{\mathbf{0}}

\newcommand{\argmin}{\operatornamewithlimits{argmin}}

\title{Deep learning for model correction of dynamical systems with data scarcity}

\author{Caroline Tatsuoka \and Dongbin Xiu\footnotemark[1]\thanks{Department of Mathematics,
		The Ohio State University, Columbus, OH 43210, USA. Emails:
		\{tatsuoka.3, xiu.16\}@osu.edu, Funding: This 
		work was partially supported by AFOSR FA9550-22-1-0011.}
				}

\begin{document}
\maketitle
\begin{abstract}
We present a deep learning framework for correcting existing dynamical system models utilizing only a scarce high-fidelity data set. In many practical situations, one has a low-fidelity model that can capture the dynamics reasonably well but lacks high resolution, due to the inherent limitation of the model and the complexity of the underlying physics. When high resolution data become available, it is natural to seek model correction to improve the resolution of the model predictions. We focus on the case when the amount of high-fidelity data is so small that most of the existing data driven modeling methods cannot be applied. In this paper, we address these challenges with a model-correction method which only requires a scarce high-fidelity data set. Our method first seeks a deep neural network (DNN) model to approximate the existing low-fidelity model. By using the scarce high-fidelity data, the method then corrects the DNN model via transfer learning (TL). After TL, an improved DNN model with high prediction accuracy to the underlying dynamics is obtained. One distinct feature of the propose method is that it does not assume a specific form of the model correction terms. Instead, it offers an inherent correction to the low-fidelity model via TL. A set of numerical examples are presented to demonstrate the effectiveness of the proposed method.
\end{abstract}

\input Introduction

\input Setup
\input Method

\input Examples

\input Conclusion

\bibliographystyle{siamplain}

\end{document}

%% file: Introduction.tex
\section{Introduction} \label{sec:intro}

 Within the field of Uncertainty Quantification (UQ), practitioners wish to account for potential sources of uncertainty in their mathematical model and, when possible, limit this uncertainty for improved model-based predictions. Model-form uncertainty has especially posed a challenge for modelers to quantify, as to do so directly would require full knowledge of the true underlying process, which in many cases is unknown \cite{UQHandbook}. Incomplete knowledge of the underlying physics leads to an inherent discrepancy between the mathematical model and the true underlying process, thus introducing error to the model’s outputs. Driven by the need for high-fidelity and low-cost models, developing model correcting techniques has been a prominent area of research within the field of UQ over the past several decades. One approach, proposed in \cite{KennedyOHagan}, attempts to correct this discrepancy by introducing an external additive correction term, which is modeled by Gaussian Process (GP). The hyper-parameters of the GP can be recovered via a maximum likelihood estimate \cite{MLEforGP} or via Bayesian inference methods \cite{KennedyOHagan, GP_params_conjugate_priors, GP_params_noninform_priors, Bayesian_MCMC_chaotic_time_series, Bayesian_MCMC_high_dim_output, Bayesian_MCMC_ANOVA}. This approach has undergone several extensions and has been successful in some applications, see \cite{PriorConstraints_KennedyOHagan, Validation_computer_models, Combining_field_data_and_computer_sims, Stats_engineering_models, Hierarchical_modeling_Qian, Bayesian_validation_Wang} and resources therein. In \cite{Xiu2016ModelCorrection}, both multiplicative and additive correction terms are introduced internally (meaning embedded within the physical processes of the model), and/or externally; these terms are then parameterized via a constrained optimization problem. The work of \cite{Multi-fid_stochastic_collocation} proposes a method in which the correction terms for a low-fidelity model are parameterized by a stochastic expansion (polynomial chaos or stochastic collocation), and an adaptive, sparse grid algorithm is used for obtaining a multi-fidelity model. In \cite{Statistical_calibration_embed} and later extended in \cite{Statistical_calibration_embed_2019}, model-error is internally embedded into the model via the model parameters, which are treated as random variables represented by polynomial chaos expansions. A Bayesian inference procedure is then suggested for parameterizing the expansion. Recently, in the context of complex dynamical systems, the work of \cite{Structural_error_complex_dynamical_systems} reviews ways to address structural error through both internal and external methods. One common feature of these approaches is that the form of the model correction, additive and/or multiplicative, needs to be chosen first. This limits the applicability of the methods because for many problems the model errors manifest themselves in much more complicated manner.

Deep learning has become a powerful tool in the approximation of unknown dynamical systems, and more recently has been used in the context of model correction in order to obtain improved system approximations. In \cite{gResNet}, the flow-map of an imperfect prior model is corrected by an additive correction term parameterized by a neural network. This deep neural network (DNN), termed the generalized residual neural network, is trained to learn the discrepancy between the flow-maps of the imperfect model and the true model. Recently, for physics-informed neural networks (PINNs), an additional deep neural network operator was proposed to learn the discrepancy between the true governing dynamics and the misspecified physics encoded in the loss function \cite{PINNs_misspecification}. See \cite{Kutz_discrepancy_model,Alzheimers_application_model_discrepancy,PINN_symbolic_model_discrepancy} for other recent works in which DNNs are used for approximating model discrepancies.

While the generalized residual network presented in \cite{gResNet} demonstrates an ability to correct imperfect dynamical model, the method requires large amounts of high-fidelity data for training the DNNs. DNNs are data hungry tools, and the performance and accuracy of their predictions typically improves as the size of the high-fidelity training data set increases. However, for many complex systems, such large-scale quantities of data may not be available, either due to the limited ability for mass data collection (common in many practical physical or biological settings), or the high computational cost for generating such data \cite{data_scarcity_article}. This introduces the challenge of data scarcity in deep learning, as the DNN cannot be sufficiently trained given the limited high-fidelity dataset. The resulting DNN model thus cannot provide accurate predictions for the underlying dynamics.

For the purposes of model correction, we propose addressing this challenge of data scarcity with a transfer learning (TL) approach. Transfer learning is a popular technique used in deep learning for handling data scarcity; transfer learning utilizes a deep neural network model trained on a large, low-fidelity training dataset to leverage the training of the model on a scarce, high-fidelity dataset, effectively `transferring' relevant model features. This tool has previously been used in the deep learning community to handle data scarcity in multi-class classification problems such as image or text classification \cite{Imagetrans,NLPtrans}. There are currently several available works which utilizes transfer learning to learn PDEs, see for example \cite{Transnet,DeepOLTransNet}. In this work, we are concerned with approximating unknown dynamical systems given scarce high-fidelity data from the true underlying dynamics. We utilize TL to correct a DNN approximation to the imperfect, “prior” model's flow-map to obtain an improved “posterior” model of the true underlying flow-map operator\footnote{While  “prior” and “posterior” are terminology borrowed from Bayesian statistics, here we strictly use prior to refer to the existing imperfect model, and posterior to refer to the corrected model.}.

Further, in many realistic settings, high-fidelity data observations may be collected on a coarse time-grid, and also may be collected over non-uniform time intervals. This poses an obstacle when the dynamics of interest lie on a finer time-scale. In \cite{inner_recurrence}, a method termed \textit{inner recurrence} is introduced to address this issue, in which the network approximation of the system's flow map is enforced on a finer time scale than the coarse data observations. In this setting, large amounts of high-fidelity data are assumed to be available to train the network, and the coarse data observations are assumed to lie on a fixed, coarse time-grid. We are able to extend this method, requiring only scarce high-fidelity data which may be collected over varying coarse time intervals in order to correct a prior model.

The remainder of the paper is organized as follows: in section 2 we present the problem statement, setup and the proposed method for model correction. In section 3, we review preliminaries, specifically recalling the flow-map learning methodology proposed in \cite{resnet} as well as the generalized residual neural network and the inner recurrence set ups proposed in \cite{gResNet} and \cite{inner_recurrence} respectively. Section 4 presents numerous numerical examples to demonstrate the effectiveness in of the method, and in section 5 we provide concluding remarks.

%% file: Setup.tex
\section{Preliminaries} 
We first present the problem setup and the related work.

\subsection{Problem Setup} \label{subsec:setup} 

Let us consider modeling an unknown autonomous dynamical system,
\begin{equation} \label{true}
          \frac{d\mathbf{x}}{dt} = \mathbf{f}(\mathbf{x}(t)), 
\end{equation}
   where $\mathbf{x} \in \mathbb{R}^n$ is the state variables and $\mathbf{f}: \mathbb{R}^n \rightarrow \mathbb{R}^n$ defines the governing equations of the system. In this setting, we assume $\mathbf{f}$ is unknown. Thus, the system cannot be solved directly for its analysis and prediction.

   We assume that we have in our possession a low-fidelity model for the system,
\begin{equation} \label{prior}
          \frac{d\widetilde{\mathbf{x}}}{dt} = \widetilde{\mathbf{f}}(\widetilde{\mathbf{x}}(t)),
\end{equation}
   where $\widetilde{\mathbf{f}}: \mathbb{R}^n \rightarrow \mathbb{R}^n$ is an approximation to the true and unknown governing equations, i.e., $\widetilde{\mathbf{f}} \approx \mathbf{f}$. This low-fidelity model may be imperfect due to simplified physics, linear approximations to nonlinear dynamics, incorrect parameterizations, etc. Throughout this paper, this low-fidelity model
   shall be referred to as the ``prior model''.
   
We also assume a small amount of high-fidelity data for the true system are available.
These high-fidelity data are scarce such that they do not allow us to directly construct a high-fidelity model of the unknown system. Our goal is to leverage the scarce high fidelity data to correct the low-fidelity prior model. The corrected model shall offer much higher predictive accuracy than the prior model. Hereafter, we shall refer to the corrected model as the ``posterior model''.

\subsection{Related Work}

Our work is related to data driven learning of unknown dynamical systems. Specifically, we utilize the framework of flow map learning (FML), as it is a flexible and
rigorous modeling approach. Although FML is a general methodology, we shall focus on its formulation in conjunction with deep neural network (DNN).

\subsubsection{Flow Map Learning}  \label{sec:FML}
For the system \eqref{true}, its flow-map is an operator $\mathbf{\Phi}: \mathbb{R}^n \times \mathbb{R} \mapsto \mathbb{R}^n$ such that $\mathbf{\Phi}(\mathbf{x}(t),s) = \mathbf{x}(t+s)$. The flow-map governs the evolution of the system over time.  Suppose one has observation data on $\mathbf{x}$ over discrete time instances separated over a constant time lag $\Delta>0$. Then the state variables over two consecutive time instances satisfies
$
       \mathbf{x}_{n+1} = \mathbf{\Phi}_\Delta(\mathbf{x}_n),
$
$\forall n$.
In FML, the goal is to conduct a numerical approximation to $\mathbf{\Phi}$. That is, find $\mathbf{N}:\mathbb{R}^n\mapsto\mathbb{R}^n$ such that $\mathbf{N}\approx\mathbf{\Phi}_\Delta$. This is accomplished by minimizing the mean squared loss
$$
\min \left\| \mathbf{x}_{n+1} - \mathbf{N}(\mathbf{x}_n)\right\|^2.
$$
When DNN is used to represent $\mathbf{N}$, the problem becomes minimization over its hyper-parameters.
From the computational point of view, it is often advantagous to utilize the ResNet structure. That is,
\begin{equation} \label{resnet}
   \mathbf{N} := \mathbf{I} + \mathcal{N}(\cdot;\mathbf{\Theta}),
\end{equation}
where $\mathbf{I}$ denotes the identity operator and $\mathcal{N}$ is a deep neural network operator with its hyper parameters $\mathbf{\Theta}$ (weights and biases of each layer). Once the DNN operator is successfully trained, the hyper parameters are fixed, and we obtain a learned model that can be used for the system prediction,
\begin{equation} \label{fml}
    \mathbf{x}_{n+1} = \mathbf{N}(\mathbf{x}_n) = \mathbf{x}_n + \mathcal{N}(\mathbf{x}_n).
\end{equation}


\subsubsection{Generalized Residual Network} \label{sec:gResNet}

The FML method using the ResNet structure \eqref{resnet} was generalized in  \cite{gResNet} as a model correction method. Assuming there exists a low fidelity prior model \eqref{prior}, its flow map operator  $\widetilde{\mathbf{\Phi}}_\Delta: \mathbb{R}^n \mapsto \mathbb{R}^n$ is then available. The work of \cite{gResNet} then proposed to find the approximate flow map to the true system \eqref{true} as
   \begin{equation} \label{gResnet}
   \mathbf{N} := \widetilde{\mathbf{\Phi}}_\Delta + \mathcal{N}(\cdot;\mathbf{\Theta}).
    \end{equation}
Compared to the ResNet structure \eqref{resnet}, it is obvious that the identify operator is replaced by the flow map operator of the prior model \eqref{prior}. The DNN operator $\mathcal{N}$ then serves as an additive correction term to the prior model.

\subsection{Our Contribution}
The contribution of this paper is in the introduction of a flexible DNN-based model correction method. The new method contains two novel features: (1) it does not assume a fixed form, either additive or multiplicative, of the model correction. Instead, the prior model is modified ``internally" by the high-fidelity data to capture the unknown dynamics. This is different from most existing model correction methods; and (2) it requires only a small set of high-fidelity data to learn the corrected posterior model. This resolves the compuational challenges of most DNN based learning methods that require a large number of high-fidelity data.



%% file: Method.tex

\section{Transfer Learning for Model Correction} 

In this section, we present the technical detail of our main method. The method is based on constructing a DNN model for the low-fidelity prior model \eqref{prior}, followed by a transfer learning re-training of the DNN model using the scarce high-fidelity data.
We also discuss how to cope with the situation when the high-fidelity data are not only
scarce in quantity but also coarse in temporal resolution. 

Throughout this paper, our discussion shall be over discrete time instances $0 \leq t_0 < t_1 < \cdots < t_k < \cdots$ over a constant time-lag $\Delta \equiv t_k - t_{k-1}$, $\forall k$.

\subsection{Scarce High-fidelity Data}

  For the unknown dynamical system \eqref{true}, we assume there are a set of trajectory data of the state variables in the following form,
   \begin{equation} \label{traj_HF}
    \left\{ \mathbf{x}(t_k^{(i)}) \right\},\qquad k = 0,1,\dots,K^{(i)}, \qquad i = 1,\dots,N,
   \end{equation}
   where $N$ is the total number of the trajectory sequences, and $K^{(i)}$ the time length of the $i$-th time sequence. We assume that these data are of high fidelity, i.e., they are highly accurate measurements of the true values of the state variables $\mathbf{x}$ at the time instances. 
   
Due to our assumption that system (2.1) is autonomous, we may omit the time index as only the relative time difference between the data observations need be noted. 
%
%
 We also reorganize the data observations into the data pairs over consecutive time instances,
   \begin{equation} \label{S-HF}
    S^{HF} \triangleq \left\{\mathbf{x}_1^{(j)},\mathbf{x}_2^{(j)} \right\}, \qquad j = 1,\dots,J_{HF},  
    \end{equation}
    where $J_{HF} = K^{(1)} + \cdots + K^{(N)}$ is the total number of such data pairs that can be extracted from the data set \eqref{traj_HF}. This is our high-fidelity training data set, which in principle shall enable one to learn the unknown dynamics of \eqref{true} via the FML framework in Section \ref{sec:FML}. However, we assume in this paper that this high-fidelity data set is scarce, in the sense that $J_{HF}$ is not sufficiently large to allow us to construct an accurate model for \eqref{true}.

    \subsection{Construction of DNN Prior Model} \label{sec:prior}

    In order to construct a high-fidelity dynamical model for the unknown system \eqref{true}, we need to utilize the low-fidelity model \eqref{prior} to supplement the scarce high-fidelity training data \eqref{S-HF}. The first step is to construct a FML model \eqref{fml} for the low-fidelity prior model \eqref{prior}. Moreover, the FML model needs to be in the form of DNN.
    
     %
     To construct the DNN prior model, we repeatedly execute the prior model \eqref{prior} to generate a large number of data pairs separated by one time lag $\Delta$. More specifically, let $\Omega\subset\mathbb{R}^n$ be the domain of interest in the phase space where we are interested in modeling the unknown system \eqref{true}.
     Let $J_{LF}\gg 1$ be the number of data pairs we seek to generate. Then, for each
     $j=1,\dots, J_{LF}$,
     \begin{itemize}
         \item Sample $\widetilde{\mathbf{x}}_1^{(j)}\in\Omega$. In most cases, random sampling with uniform distribution in $\Omega$ is sufficient;
         \item Solve the prior model \eqref{prior} with an initial condition $\widetilde{\mathbf{x}}_1^{(j)}$  over one time step $\Delta$ to
         obtain $\widetilde{\mathbf{x}}_2^{(j)}= \widetilde{\mathbf{\Phi}}_\Delta\left(\widetilde{\mathbf{x}}_1^{(j)}\right)$.
     \end{itemize}
    Upon conducting these simulations, we obtain a low-fidelity training data set,
   \begin{equation} \label{S-LF}
    S^{LF} \triangleq \left\{\widetilde{\mathbf{x}}_1^{(j)},\widetilde{\mathbf{x}}_2^{(j)} \right\}, \qquad j = 1,\dots, J_{LF}.  
\end{equation}
This procedure should not present a significant computational challenge,
assuming the low-fidelity model \eqref{prior} can be solved fast and inexpensively. 

We then construct a FML model based on this low-fidelity data set using DNN, in the form of \eqref{resnet}. Let $\widetilde{\mathbf{N}}(\cdot; \Theta):\mathbb{R}^n \rightarrow \mathbb{R}^n$ be the
corresponding DNN operator with its hyper parameters $\Theta$. The learning of $\widetilde{\mathbf{N}}$ is accomplished by minimizing the mean squared loss over the low-fidelity data set \eqref{S-LF}, i.e.,
\begin{equation} \label{Theta}
\Theta^* = \argmin_{\Theta} \sum_{j=1}^{J_{LF}} \left\| \widetilde{\mathbf{x}}_2^{(j)} - \widetilde{\mathbf{N}}\left(\widetilde{\mathbf{x}}_1^{(j)};\Theta\right)\right\|^2.
\end{equation}
Once the optimization is finished, the hyper parameters $\Theta^*$ are fixed and shall be suppressed in our exposition, unless confusion arises otherwise.

To facilitate the discussion, it is necessary to examine the detail of $\widetilde{\mathbf{N}}$. Both the input layer and output layer have $n$ nodes, corresponding to the dimension of the state variable $\mathbf{x}$.
   Let us assume we have $M\geq 1$ hidden layers, each of which contains $d\geq 1$ nodes. 
   The DNN operator can be written as
the following composition of affine and nonlinear transformations,
    \begin{equation} \label{operator}
        \widetilde{\mathbf{N}} = \mathbf{W}_M \circ \left(\sigma_{M} \circ {\mathbf{W}}_{M-1}\right) \circ \cdots  \circ \left({\sigma}_1 \circ {{\mathbf{W}}}_0\right),
    \end{equation}
    where ${\mathbf{W}}_i$ is the weight matrix containing the weights connecting the $i^{th}$ and $(i+1)^{th}$ layer and the biases in the $(i+1)^{th}$ layer, $\sigma_i$ is the (nonlinear) component-wise activation function of the $i^{th}$ layer, and the $0^{th}$ layer is the input layer. 
    At the output layer, the $(M+1)^{th}$ layer, the linear activation function ($\sigma(x) = x$) is used. This results in an identity operator, which is suppressed.

Let us define a shorthanded notation
\begin{equation} \label{W}
\mathbf{W}_{[0:m]} = [\mathbf{W}_0, \dots, \mathbf{W}_m], \qquad 0\leq m\leq M.
\end{equation}
The hyper parameters $\Theta$ then refer to the collection of all the weight matrices, i.e.,
$\Theta = \mathbf{W}_{[0:M]}$. 
The parameter optimization problem \eqref{Theta} can be written equivalently as
\begin{equation} \label{prior_loss}
    \widetilde{\mathbf{W}}^*_{[0:M]} = \argmin_{\mathbf{W}_{[0:M]}}  \sum_{j=1}^{J_{LF}} \left\| \widetilde{\mathbf{x}}^{(j)}_2 - \widetilde{\mathbf{N}}(\widetilde{\mathbf{x}}^{(j)}_1; \mathbf{W}_{[0:M]}) \right\|^2. 
\end{equation}
Once the training is completed, we obtain the DNN prior model
    \begin{equation} \label{DNN_prior}
        \widetilde{\mathbf{x}}_{n+1} = \widetilde{\mathbf{N}}(\widetilde{\mathbf{x}}_n; \widetilde{\mathbf{W}}^*_{[0:M]}).
    \end{equation}
    Note that this DNN prior model is, at best, as accurate as the prior low-fidelity model \eqref{prior}.
    We remark that if the low-fidelity model \eqref{prior} is already in the form of a DNN, this step can be avoided as the prior DNN model already exists. 

\subsection{Transfer Learning for Model Correction} \label{subsec:sec3.3}

The basic premise is that the trained DNN prior model \eqref{DNN_prior}, which is an accurate representation of the prior model \eqref{prior}, is able to capture the ``bulk" behavior of the dynamics of the unknown system \eqref{true}. To further improve/correct the DNN prior model, we employ transfer learning (TL) technique, with the help of the scarce high-fidelity data set \eqref{S-HF}.

The principle of transfer learning (TL) is based on the widely accepted notation that the early layers of a DNN extract more general features of a dataset, while later layers contain higher-level features (\cite{transfer_learning_neyshabur,transfer_learning_yosinski,transfer_learning_raghu}).
Following this, we ``freeze" the majority of the layers in the DNN prior model \eqref{DNN_prior}. Specifically, we fix the weights and biases in most of the layers of the trained DNN prior model by making them un-modifiable. We then use the high-fidelity data set \eqref{S-HF} to retrain the parameters in the last few layers.

Let $0\leq \ell\leq M$ be a number separating the layers in the DNN operator $\widetilde{\mathbf{N}}$ of the trained prior model \eqref{DNN_prior} into two groups: the first $\ell$ layers from the input layer ($0^{th}$ layer) to the $(\ell-1)^{th}$ layer, and the second group from the $\ell^{th}$ layer to the output layer ($M^{th}$ layer).
Using the notation \eqref{W}, the hyper parameters can be 
separated into the following two groups correspondingly,
\begin{equation} \label{W_split}
    \mathbf{W}_{[0:M]} = \left[\mathbf{W}_{[0:\ell-1]}, \mathbf{W}_{[\ell:M]}\right].
\end{equation}

We fix the first group of parameters to be at the values trained in the DNN prior model via \eqref{prior_loss}, i.e. $\mathbf{W}_{[0:\ell-1]} = \widetilde{\mathbf{W}}^*_{[0:\ell-1]}$, and re-train the second group of parameters by minimizing the mean square error of the model \eqref{DNN_prior} over the high-fidelity data set \eqref{S-HF}.
%
 The optimization problem then becomes
    \begin{equation} \label{TL-loss}
   \mathbf{W}^*_{[\ell:M]} = \argmin_{\mathbf{W}_{[\ell:M]}}   \sum_{j=1}^{J_{HF}} \left\| \mathbf{x}^{(j)}_2 - \widetilde{\mathbf{N}}\left(\mathbf{x}^{(j)}_1; \widetilde{\mathbf{W}}^*_{[0:\ell-1]}, \mathbf{W}_{[\ell:M]}\right) \right\|^2.
\end{equation}
Once training is completed, we obtain a DNN whose hyper-parameters are
\begin{equation} \label{W_final}
\mathbf{W}^*_{[0:M]} =\left[ \widetilde{\mathbf{W}}^*_{[0:\ell-1]}, \mathbf{W}^*_{[\ell:M]}\right],
\end{equation}
where the first group is from \eqref{prior_loss} and the second group from \eqref{TL-loss} separately.

Finally, we define our posterior DNN model operator $\mathbf{N}$ using these parameters \eqref{W_final}, i.e.,
\begin{equation} \label{DNN-post}
    \mathbf{N} \triangleq \widetilde{\mathbf{N}}\left(\cdot; \mathbf{W}^*_{[0:M]}\right).
\end{equation}
Subsequently, we obtain a predictive model using the posterior DNN: for any given initial condition $\mathbf{x}_0$,
    \begin{equation} \label{DNN_post}
        {\mathbf{x}}_{n+1} = {\mathbf{N}}({\mathbf{x}}_n), \qquad n=0,1,\dots,
    \end{equation}
    
In most applications, TL only needs to modify the last few layers of a pre-trained DNN. In our setting, this means $\ell\approx M$. Therefore, the total number of parameters to be re-trained in \eqref{TL-loss} is relatively small in comparison to the total number of network parameters.
This is highly relevant to the scarce high-fidelity data case we consider here. In fact, for many problems, one may take $\ell=M$, which implies that TL only needs to re-train the parameters in the output layer.

\subsection{Connection with Least Squares}

Let us consider the special case of $\ell=M$, when TL only re-trains the output layer. The parameter grouping \eqref{W_split} becomes
$$
\mathbf{W}_{[0:M]} = \left[\mathbf{W}_{[0:M-1]}, \mathbf{W}_{M}\right],
$$
 and the minimization problem \eqref{TL-loss} become 
 \begin{equation} \label{MSE}
   \mathbf{W}^*_{M} = \argmin_{\mathbf{W}_{M}}   \sum_{j=1}^{J_{HF}} \left\| \mathbf{x}^{(j)}_2 - \widetilde{\mathbf{N}}\left(\mathbf{x}^{(j)}_1; \widetilde{\mathbf{W}}^*_{[0:M-1]}, \mathbf{W}_{M}\right) \right\|^2,
\end{equation} 
where $\widetilde{\mathbf{W}}^*_{[0:M-1]}$ are the fixed parameters pre-trained via \eqref{prior_loss}.
Since the linear activation function $\sigma(x)=x$ is used in the output layer, this minimization problem becomes a least squares problem,
   \begin{equation} \label{LSQ}
       \mathbf{A}\mathbf{W}_{M} = \mathbf{B},
   \end{equation}
  where the matrix $\mathbf{A}$ denotes the ``feature matrix" of DNN prior model and $\mathbf{B}$ is the high-fidelity data matrix.

To see this relation clearly, let us consider the connection from the $(M-1)^{th}$ layer, which is the last hidden layer with $d\geq 1$ nodes, to the output layer which has $n\geq 1$ nodes. For each $i=1,\dots, n$, consider the $i^{th}$ node of the output layer. Let $\mathbf{w}_{M}^i \in \mathbb{R}^d$ be the weights from the $d$ nodes of the $(M-1)^{th}$ layer, and $b_M^i$ be its scalar bias term. Following the standard notation, we concatenate the weights and bias into a single vector $\bm{\beta}_i = [b_{M}^i; \mathbf{w}_{M}^i] \in \mathbb{R}^{d+1}$. The matrix $\mathbf{W}_M$ thus takes the form
\begin{equation} \label{WM}
    \mathbf{W}_M = \left[\bm{\beta}_1, \cdots, \bm{\beta}_n\right]
\in \mathbb{R}^{(d+1) \times n}.
\end{equation}

For the feature matrix $\mathbf{A}$, consider the operator of the DNN prior model \eqref{operator} and let $\mathbf{a}:\mathbb{R}^n\to\mathbb{R}^d$ be the output function of the $(M-1)^{th}$ layer,
$$
   \mathbf{a}(\mathbf{x}) = \left({\sigma}_{M} \circ \widetilde{{\mathbf{W}}}^*_{M-1}\right) \circ \cdots \circ \left({\sigma}_1 \circ \widetilde{{\mathbf{W}}}^*_0\right)(\mathbf{x}).
$$
Note this is a fixed function as the parameters $\widetilde{\mathbf{W}}^*_{[0:M-1]}$ are pre-trained via \eqref{prior_loss} during the DNN prior model construction.
Let
$$
\hat{\mathbf{a}}(\mathbf{x}) = [1; \mathbf{a}(\mathbf{x})] \in \mathbb{R}^{d+1},
$$
where the inclusion of $1$ is to accommodate the bias term in the definition of $\mathbf{W}_M$ \eqref{WM}.
The feature matrix is then formed by evaluating the output vectors at each of the first entry of the high-fidelity data pairs \eqref{S-HF}, i.e.,
$$
\mathbf{A} = \left[\hat{\mathbf{a}}\left(\mathbf{x}^{(1)}_1\right)^T;\cdots;
\hat{\mathbf{a}}\left(\mathbf{x}_1^{(J_{HF})}\right)^T\right] \in \mathbb{R}^{J_{HF} \times (d+1)}.
$$

Finally, the matrix $\mathbf{B}$ contains the second entries of the high-fidelity data \eqref{S-HF},
$$
\mathbf{B} = \left[ \left(\mathbf{x}^{(1)}_2\right)^T; \cdots, \left(\mathbf{x}^{(J_{HF})}_2\right)^T \right]
\in \mathbb{R}^{J_{HF} \times n}.
$$
%

Upon solving the least squares problem \eqref{LSQ} and obtaining its solution $\mathbf{W}^*_{M}$, we obtain the hyper parameters for the posterior model
\begin{equation}
\mathbf{W}^*_{[0:M]} =\left[ \widetilde{\mathbf{W}}^*_{[0:M-1]}, \mathbf{W}^*_{M}\right].
\end{equation}
The corresponding DNN operator \eqref{DNN_post} is
\begin{equation} \label{post_oper}
    {\mathbf{N}} = \mathbf{W}^*_M \circ \left(\sigma_{M} \circ \widetilde{\mathbf{W}}^*_{M-1}\right) \circ \cdots  \circ \left({\sigma}_1 \circ \widetilde{{\mathbf{W}}}^*_0\right).
\end{equation}
It is obvious that the presented method can be viewed as an ``internal" correction method to the DNN prior model. This is different from most of the existing model correction methods, where certain forms of the correction (additive or multiplicative) are assumed.



%

\subsection{High-fidelity Data Over Coarse Time Scale} \label{sec:coarse}

We now discuss a situation when the high-fidelity data are not only scarce in quantity, they are also acquired over a time scale larger than desired. This is a rather common situation, as high-fidelity data are often difficult to acquire freely. 

Specifically, let $\delta>0$ be a time step over which the true system \eqref{true} can be sufficiently analyzed and studied. Suppose the high-fidelity trajectory data \eqref{traj_HF} are already formed into the pairwise data set \eqref{S-HF}, where each pair has a time step (much) bigger than $\delta$. Denote the high-fidelity dataset as
    \begin{equation} \label{coarse_dataset} 
   S^{HF} = \left\{\mathbf{x}^{(j)}_1,{\mathbf{x}}^{(j)}_2; \Delta^{(j)}\right\}, \quad j = 1,\dots,{J_{HF}}.  
   \end{equation}
where $\Delta^{(j)}>\delta$ is the time step separating the pair. For simplicity, we assume $\Delta^{(j)} = k^{(j)}\cdot \delta$ with $k^{(j)}\geq 1$ is an integer for all $j$. Our objective is to create an accurate high-fidelity FML model for the unknown system \eqref{true} over the desired  time step $\delta$.


To accomplish this, we first train the DNN prior model over the required time step $\delta$ using the procedure described in Section \ref{sec:prior}. Specifically, we
\begin{itemize}
    \item Generate the low-fidelity data set \eqref{S-LF} using the time step $\delta$, i.e., $\widetilde{\mathbf{x}}_2^{(j)}= \widetilde{\mathbf{\Phi}}_\delta\left(\widetilde{\mathbf{x}}_1^{(j)}\right)$, $j=1,\dots, J_{LF}$;
    \item Train the DNN via \eqref{prior_loss};
    \item Obtain DNN prior model
      \begin{equation} \label{new_prior}
        \widetilde{\mathbf{x}}_{n+1} = \widetilde{\mathbf{N}}_\delta (\widetilde{\mathbf{x}}_n; \widetilde{\mathbf{W}}^*_{[0:M]}),
    \end{equation}
    where the subscript $\delta$ is added to emphasize the model is over the small time step.
\end{itemize}
%
%
%
%

Transfer learning model correction is then conducted by utilizing the coarse high-fidelity dataset \eqref{coarse_dataset}. 
Since each $\Delta^{(j)} = k^{(j)}\cdot \delta$ is a multiple of the small step $\delta$, we modify the TL training \eqref{TL-loss} using $k^{(j)}$ composition of the DNN prior operator. 
Let
$$
\widetilde{\mathbf{N}}_\delta^{[k]} = \widetilde{\mathbf{N}}_\delta\circ\cdots\circ\widetilde{\mathbf{N}}_\delta, \qquad k\textrm{ times},
$$
be the $k$ times composition of $\widetilde{\mathbf{N}}_\delta$. We conduct
the transfer learning via
%
\begin{equation} \label{TL-loss-recurrent-loss}
\mathbf{W}^*_{[\ell:M]} = \argmin_{\mathbf{W}_{[\ell:M]}}   \sum_{j=1}^{J_{HF}} \left\| \mathbf{x}^{(j)}_2 - \widetilde{\mathbf{N}}_\delta^{[k^{(j)}]}\left(\mathbf{x}^{(j)}_1; \widetilde{\mathbf{W}}^*_{[0:\ell-1]}, \mathbf{W}_{[\ell:M]}\right) \right\|^2.
\end{equation}


Once sufficiently trained, all the hyper parameters are fixed in the form of \eqref{W_final}, and we obtain 
the posterior DNN model
\begin{equation} \label{post_delta}
    \mathbf{N}_\delta \triangleq \widetilde{\mathbf{N}}_\delta \left(\cdot; \mathbf{W}^*_{[0:M]}\right).
\end{equation}
Note that the posterior model is now defined over the required small time step $\delta$.
The use of the recurrent loss \eqref{TL-loss-recurrent-loss} follows from the work of \cite{inner_recurrence}, where the mathematical justification of the approach was discussed.

%% file: Examples.tex

\section{Computational Studies} \label{sec:examples} We now present numerical studies to demonstrate the effectiveness of the proposed method. The numerical examples are organized in two groups. In the first group, the scarce high-fidelity data are observed on the same time step required by the system resolution; whereas in the second group they are observed over a coarser time step, thus requiring the technique discussed in Section \ref{sec:coarse}.
In all the examples, the true models are known. They are used only to generate the high-fidelity training data, as well as the validation data to examine the performance of the trained DNN posterior models.

%
%

\subsection{Model Correction with Scarce Data} 
\label{subsec:sec4.1}

Here we assume the scarce high-fidelity data \eqref{S-HF} are observed over a time step $\Delta$ that is sufficient to resolve the unknown dynamics \eqref{true}. In each of the examples here, a low-fidelity prior model is known and used to construct the DNN prior model. The DNNs have $3$ hidden layers, while the number of neurons per layer varies from $20$ to $80$. 
The number of training data $J_{LF}$ \eqref{S-LF} for the DNN prior model is typically  $\sim 5$ times the number of network parameters. 
The DNN prior model is first trained for $10,000 - 20,000$ epochs using a batch size of $100$. Transfer learning is then conducted on the last layer $\mathbf{W}_M$ for model correction with the high-fidelity data set \eqref{S-HF}, where the number of data $J_{HF}$ is $60\sim 200$ times smaller than that of the low-fidelity data \eqref{S-LF}. The Adam optimizer with a learning rate of $1e-3$ is used for training.

\subsubsection{Damped Pendulum} 
Let us consider a damped pendulum as the unknown true system,
    \begin{equation} \label{damp}
    \left\{
    \begin{split}
         \dot{x}_1 & = x_2\\
        \dot{x}_2 & = -\alpha x_2 - \beta \sin(x_1), 
    \end{split}
    \right.
    \end{equation}
where $\alpha = 0.1$ and $\beta = 9$.    
Our prior model is the following harmonic oscillator, 
   \begin{equation} \label{osc}
   \left\{
    \begin{split}
         \dot{x}_1 & = x_2\\
        \dot{x}_2 & = -\beta x_1. 
    \end{split}
    \right.
    \end{equation}
Although a very simple system, this represents the typical case of using linearization to model a nonlinear system.

\begin{figure}
\centering
  \includegraphics[width=0.49\linewidth]{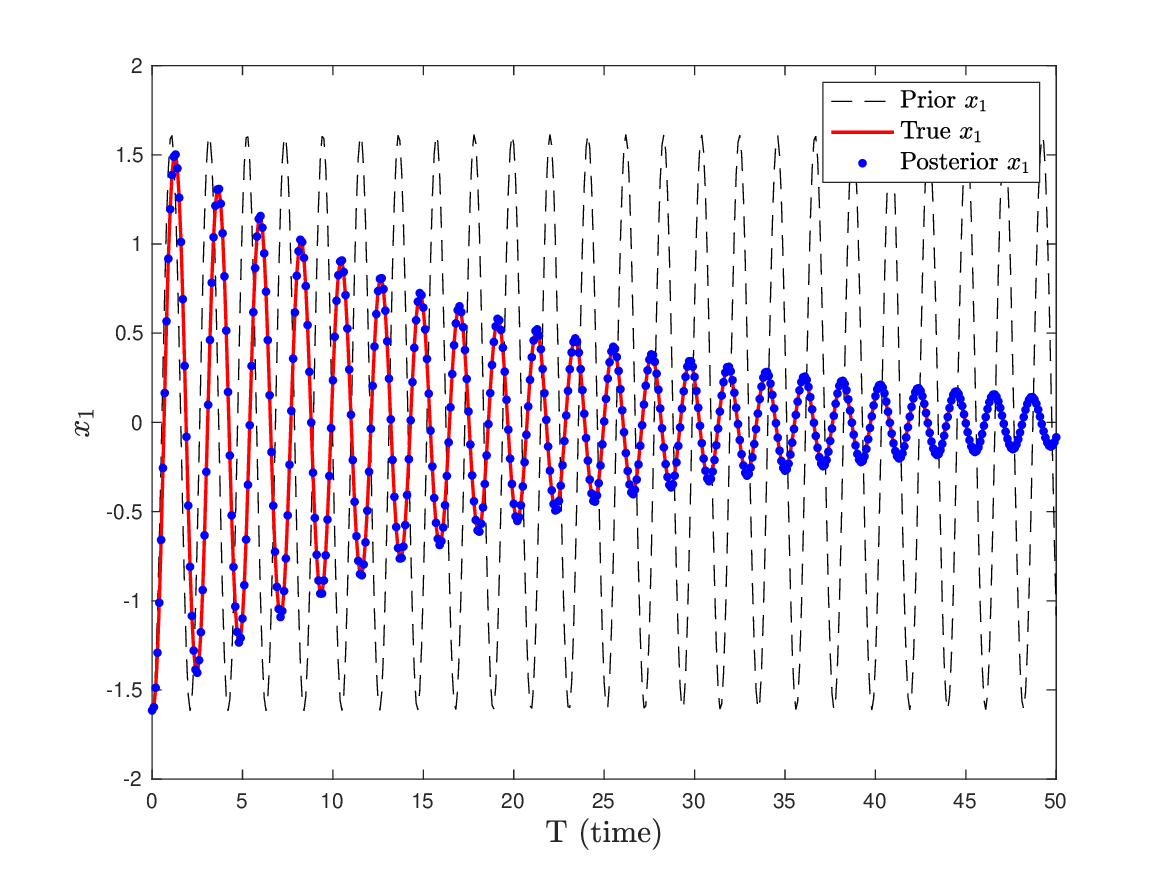}
  \includegraphics[width=0.49\linewidth]{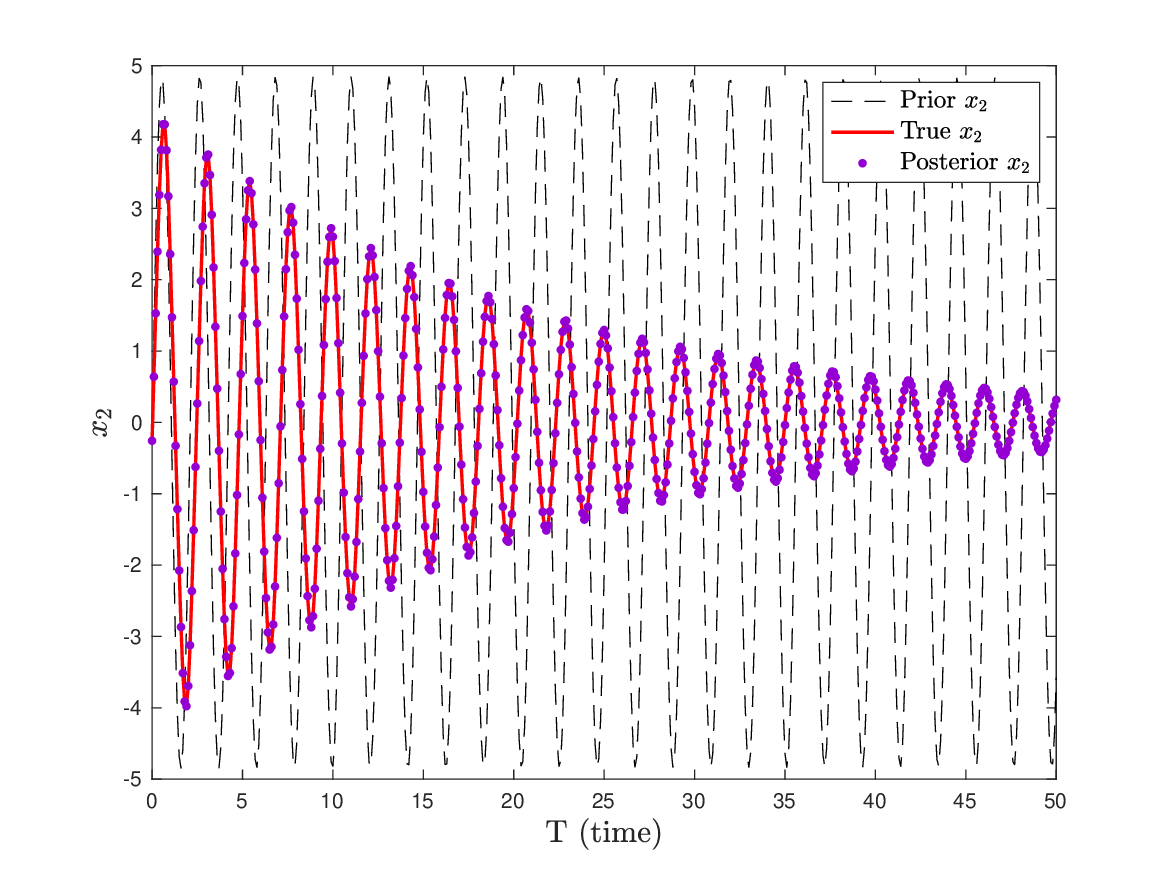}
\caption{Damped pendulum: system prediction up to $T=50$. Left: $x_1$; Right: $x_2$.}
\label{damped_x}
\end{figure}

\begin{figure}
    \centering \includegraphics[width=0.75\linewidth]{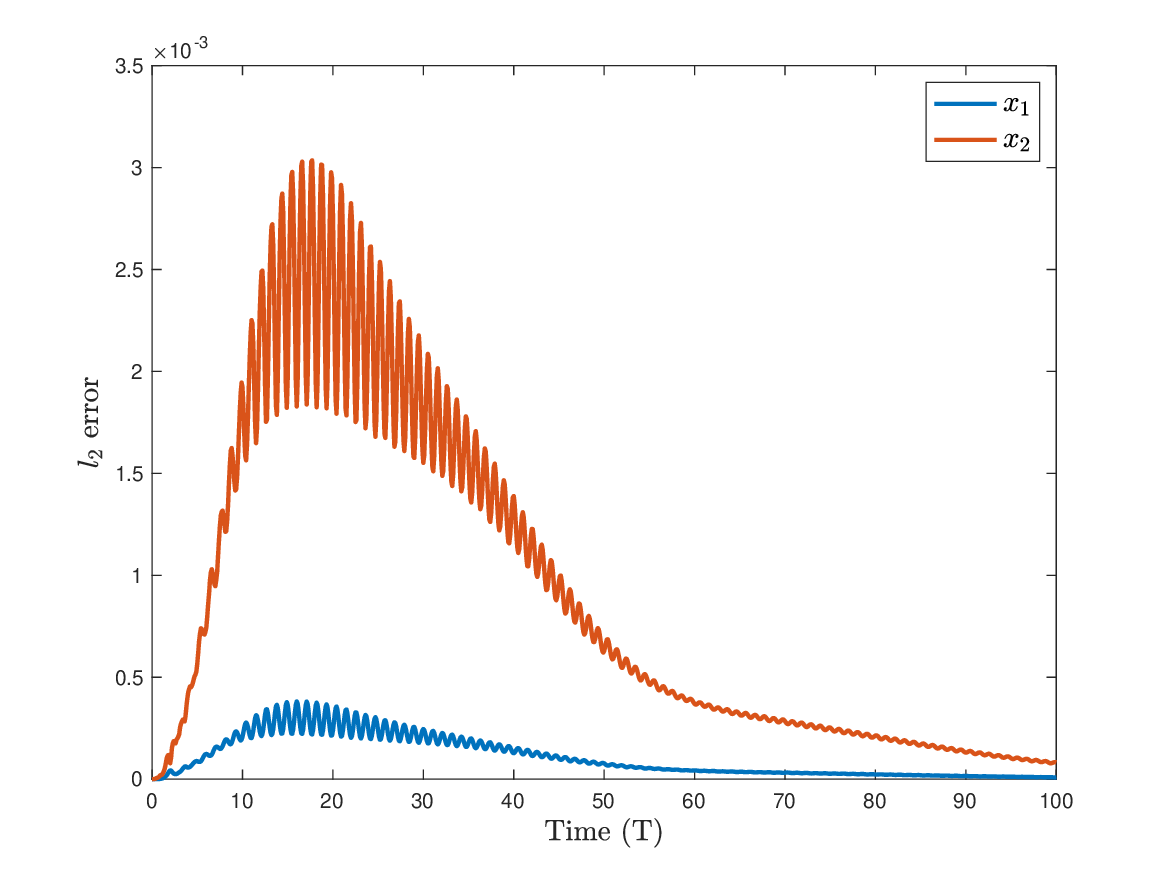}
    \caption{Damped pendulum: prediction errors up to $T=100$.}
    \label{damped_error}
\end{figure}

We consider the modeling domain $\Omega = [-\pi,\pi] \times [-2\pi,2\pi]$ and use a constant time step $\Delta = 0.1$. Data pairs are collected by randomly sampling initial conditions from $\Omega$ and evolving the state variables over one time step. The number of nodes per layer in our DNN is taken to be 50. The DNN model is trained for 10,000 epochs with 30,000 low-fidelity data pairs generated from \eqref{osc}. We then use 250 high-fidelity data pairs from the true system \eqref{damp} to correct the DNN prior model. Figure \ref{damped_x} shows an example trajectory prediction with a randomly selected initial condition $\mathbf{x}_0=(-1.615, -0.258)$. We can clearly observe that the prior model fundamentally mis-characterizes the system behavior. With the small high-fidelity data set, the posterior model can correct the prior model and accurately predict the system behavior. Figure \ref{damped_error} shows the errors in the prediction for up to $T=100$, averaged over 100 trajectories. We observe good accuracy in the posterior model prediction for such a relatively long term.

\subsubsection{Duffing Equation} 

We consider the Duffing equations as our true unknown model,
\begin{equation} \label{duffing}
\left\{
\begin{split}  
    \dot{x_1} &= x_2 \\
    \dot{x_2} &= -x_1 -\epsilon x_1^{3},
 \end{split}
 \right.
\end{equation}
where $\epsilon = 0.05$. The prior low-fidelity model is again the linear harmonic oscillator \eqref{osc}.

The modeling domain is taken to be $\Omega = [0,3]^2$. Using prior model \eqref{osc}, a total of $30,000$ data pairs over a time lag $\Delta = 0.1$ are randomly distributed within the time frame $T=12$ to train the DNN. The DNN model is  taken to have 50 neurons per layer and is trained for 10,000 epochs.

\begin{figure}

\centering
  \includegraphics[width=0.49\linewidth]{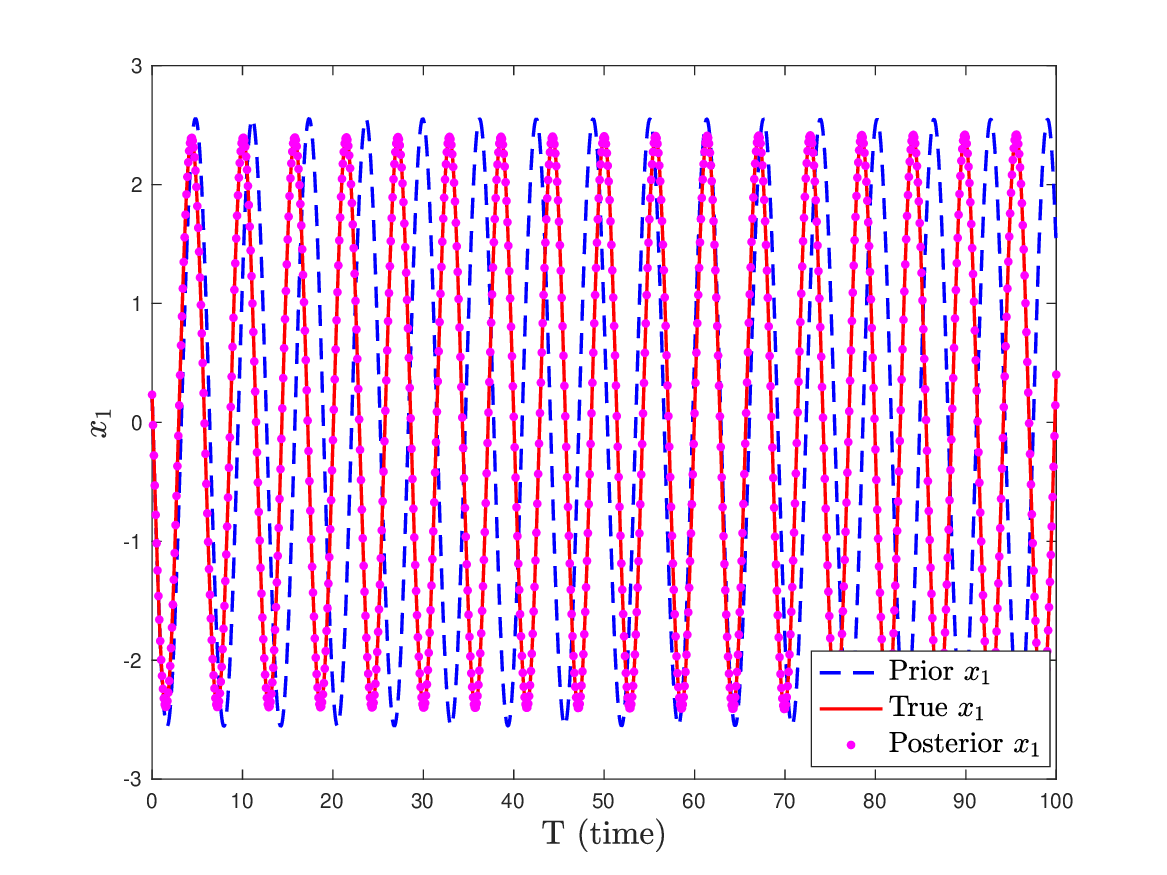}
  \includegraphics[width=0.49\linewidth]{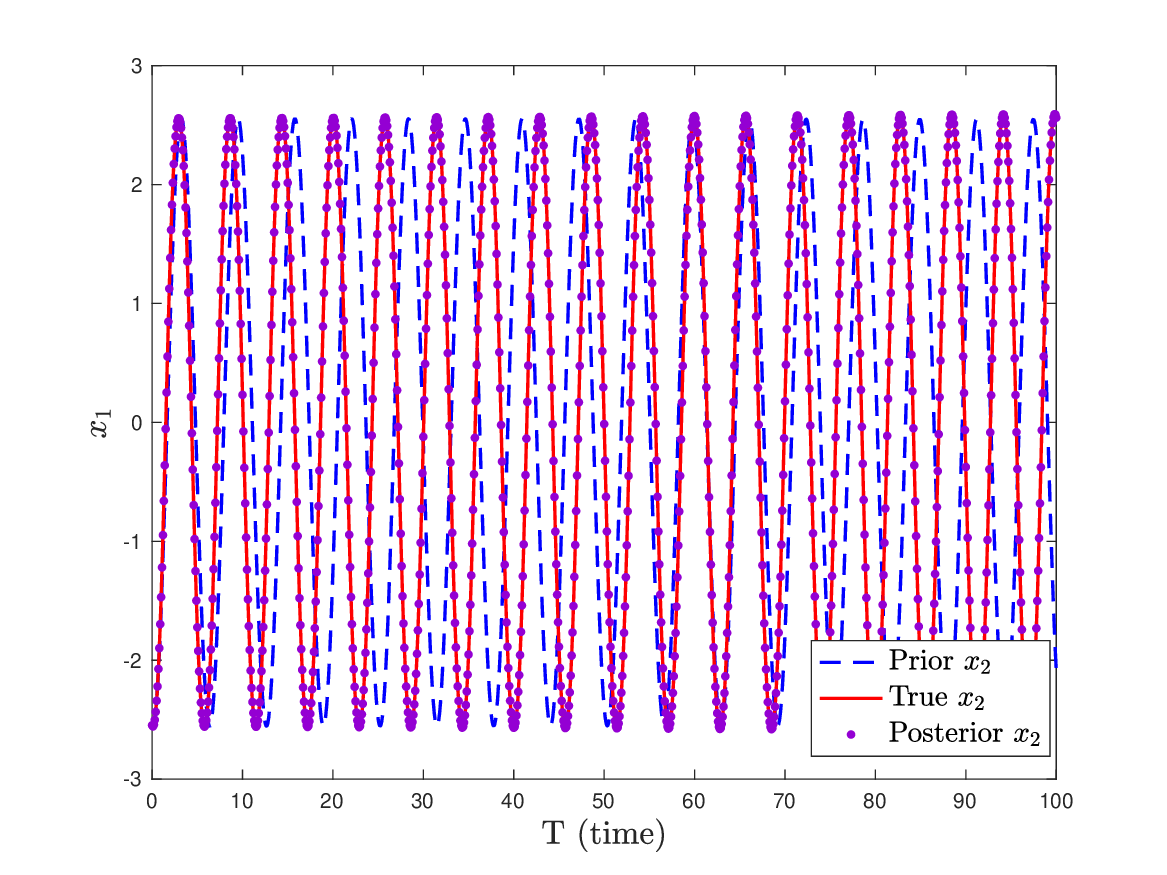}
  \caption{Duffing equation: system prediction up to $T=100$. Left: $x_1$; Right: $x_2$.}

  \label{fig:duffing}
\end{figure}

\begin{figure}
    \centering \includegraphics[width=0.75\linewidth]{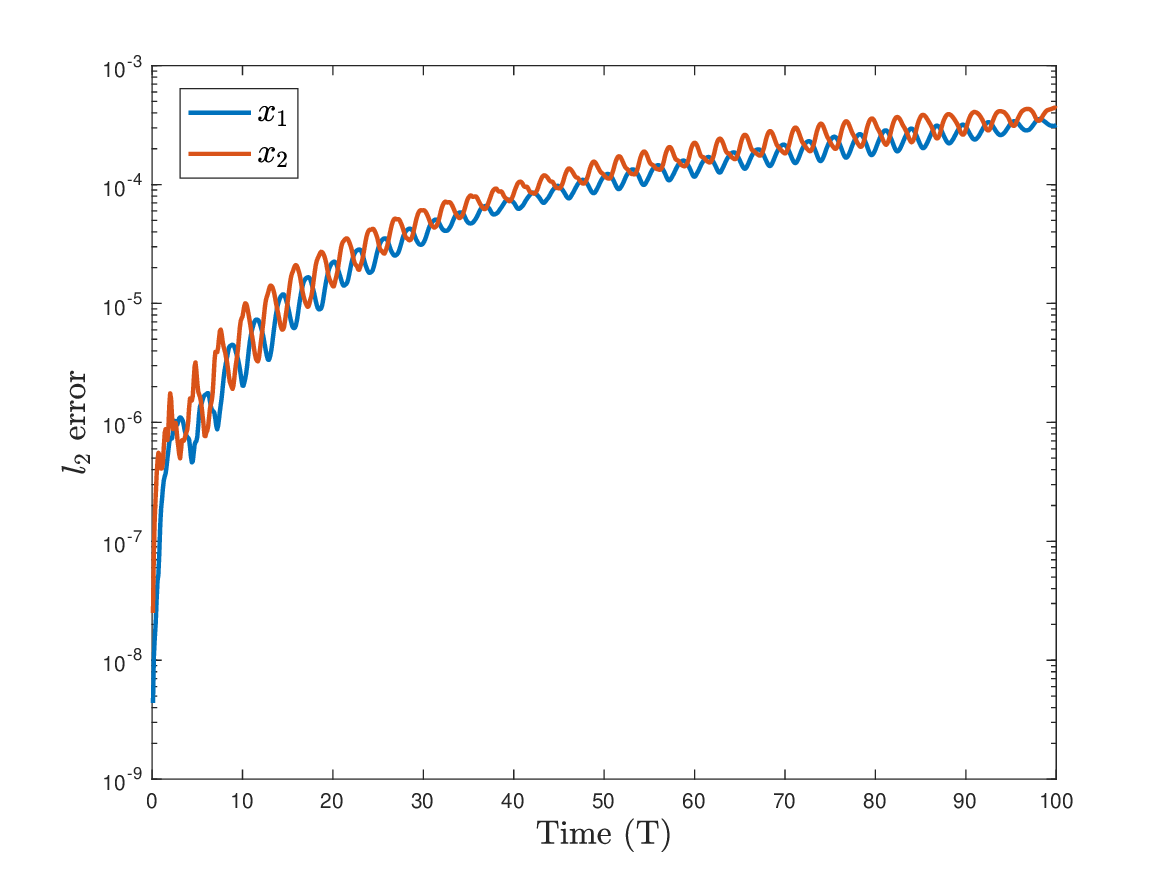}
    \caption{Duffing equation: average $l_2$ error over 100 test trajectories.}
    \label{duffing_error}
\end{figure}

A total of $500$ high-fidelity data pairs are generated from the true system \eqref{duffing} for model correction. For validation results, we present a case using a randomly selected initial condition $\mathbf{x}_0 = (0.233, -2.547)$ in Figure \ref{fig:duffing}. For this relatively long-term prediction up to $T=100$, we observe that the posterior model is able to correct the deficiency in the low-fidelity prior model and accurately predict the system behavior.   The prediction error, averaged over 100 randomly selected trajectories, is shown in Figure \ref{duffing_error}. 
We observe good accuracy from the posterior model.  

\subsubsection{SEIR Model} 

We now consider the true unknown system to be a SEIR model,
\begin{equation} \label{seir}
\left\{
\begin{split}
       	\dot{S} &= {\mu}(1-S) - {\beta} SI, \\
		\dot{E} &= {\beta} SI - ({\mu} +{\sigma})E, \\ 
            \dot{I} &= {\sigma} I - ({\mu} + {\gamma})I, \\ 
            \dot{R} &= {\gamma} I - {\mu} R, 
\end{split}
\right.
\end{equation}
where the susceptible (S), exposed (E), infected (I), and recovered (R) are the populations in a closed epidemic. They are normalized against the total population  such that $S,E,I,R \in [0,1]^4$ and $S + E + I + R = 1$ at all times $t$. We assume that the unknown aspect of the system is in the parameters: the true values of the parameters lie within the following ranges: $\mu \in [0.1,0.5]$, $\beta \in [0.7, 0.11]$, $\sigma \in [0.3,0.7]$, and $\gamma \in [0,0.4]$. 

For the prior model, we consider the same SEIR system \eqref{seir} with incorrect parameter values. Specifically, we assume in the prior model all parameters are fixed at the mean values of their corresponding ranges, i.e.,  $\mu = 0.3, \beta = 0.9, \sigma = 0.5, \gamma = 0.2$. We generate 30,000 low-fidelity data pairs over a time lag $\Delta = 0.2$ to train our DNN model, and data pairs are randomly generated from trajectories of length $T=5$. The number of nodes per layer is 50. 

\begin{figure}
\centering
\includegraphics[width = \textwidth]{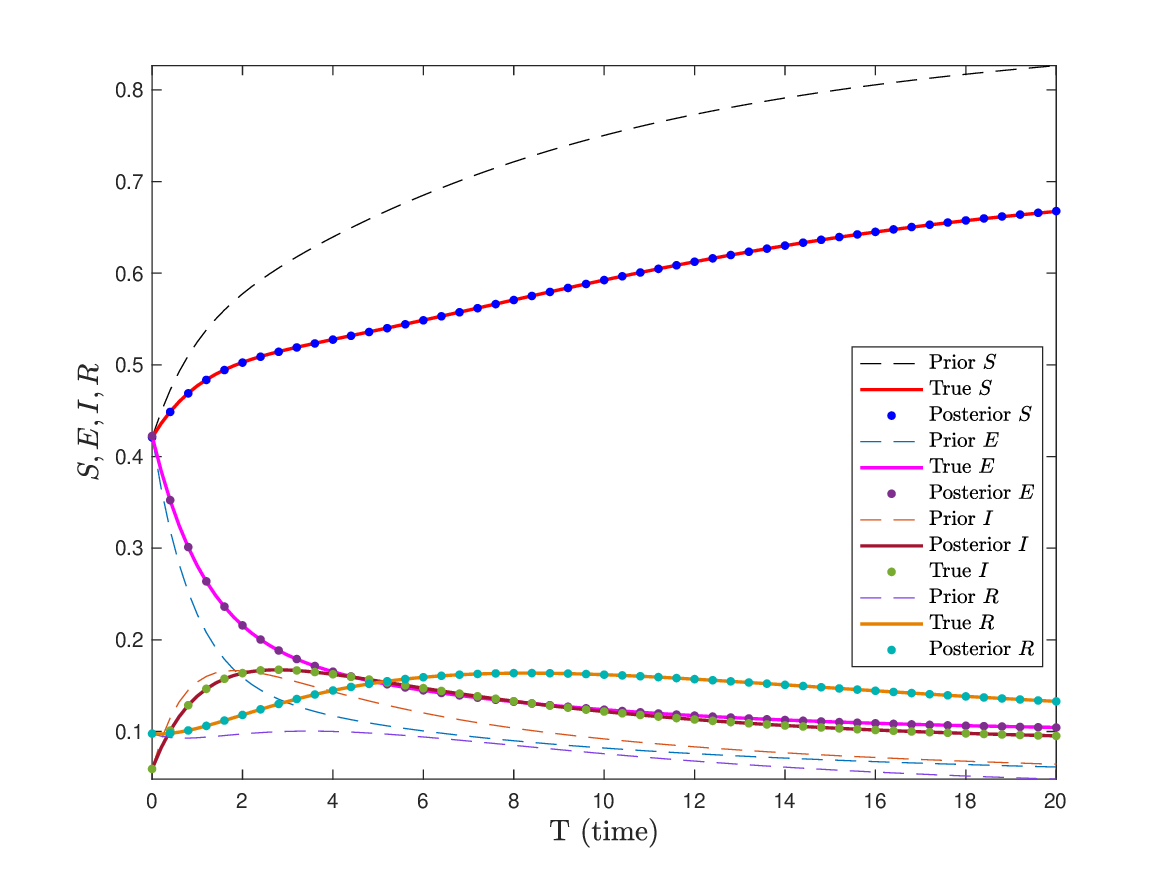}
    \caption{SEIR Model: system prediction up to $T=20$.} 
    \label{SEIR_example}
\end{figure}

\begin{figure}
\centering
\includegraphics[width = \textwidth]{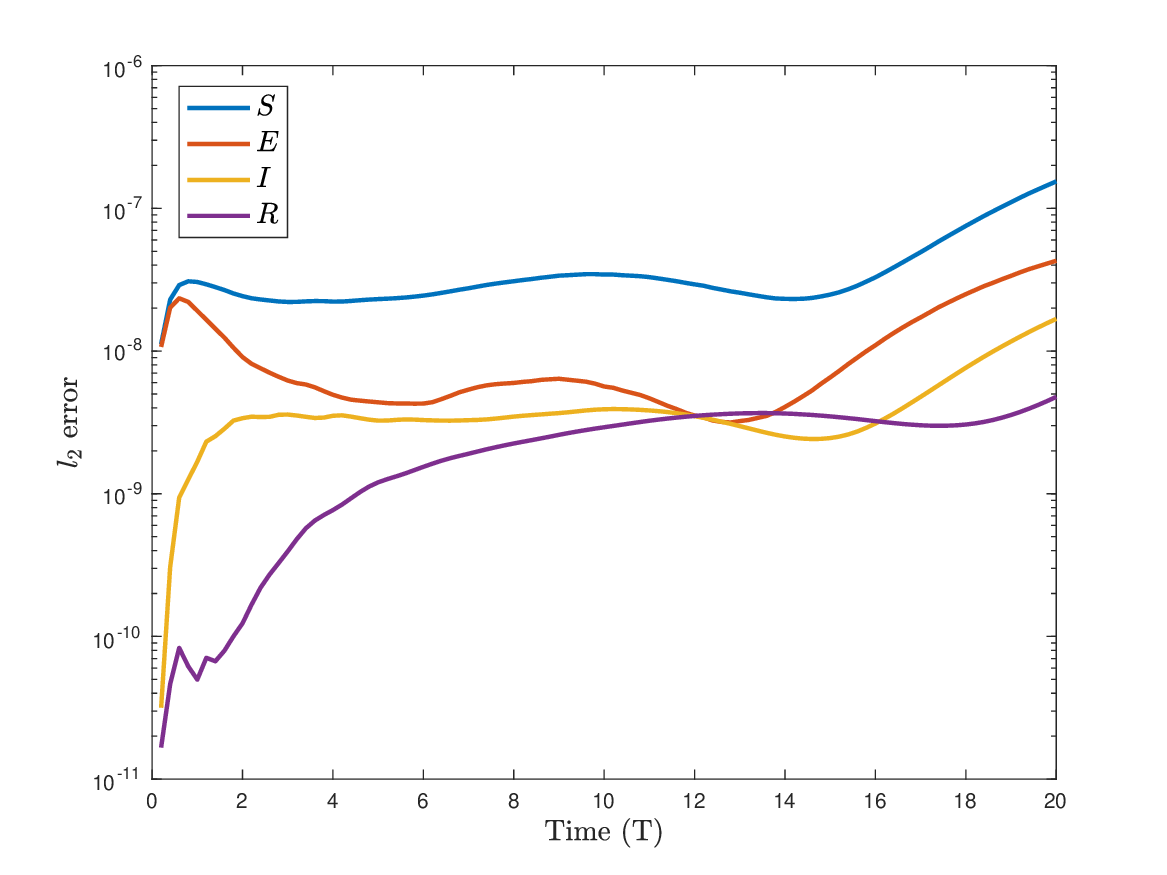}
    \caption{SEIR model, average $l_2$ error over 100 test trajectories.}
    \label{SEIR_error}
\end{figure}

For model correction, we assume the true model is the system \eqref{seir} with the parameters $\mu = 0.1792$, $\beta=0.8669$, $\sigma = 0.3562$, and $\gamma = 0.2235$ (note these values remain unknown to the DNN models). We generate 250 high-fidelity data pairs from the true model for the model correction procedure. In Figure \ref{SEIR_example} we plot solution trajectories with for the randomly selected initial condition Initial condition $\mathbf{x}_0 = (0.4208,0.4224,0.0592,0.0976)$. While the prior model, with its incorrect parameters, shows significant errors in the prediction, the posterior model is able to correct the errors and accurately capture the system behavior.
The high accuracy of the posterior model can be seen in the error plot Figure \ref{SEIR_error}.

\subsubsection{Metabolic Pathway} 

We now examine the three-step metabolic pathway example \cite{engl2009inverse}, a nonlinear system of eight equations describing the conversion of a substrate to product via intermediate metabolites $M_1$ and $M_2$, enzymes $E_1, E_2, E_3$ and mRNAs $G_1, G_2, G_3$:
\begin{equation}
\left\{
\begin{split}
    \frac{d G_1}{d t} &= \frac{V_1}{1 + (\frac{P}{K_{i1}})^{n_{i1}} + (\frac{K_{a1}}{S})^{n_{a1}}} - k_1 \cdot G_1 \\
    \frac{d G_2}{d t} &= \frac{V_2}{1 + (\frac{P}{K_{i2}})^{n_{i2}} + (\frac{K_{a2}}{M_1})^{n_{a2}}} - k_2 \cdot G_2 \\
    \frac{d G_3}{d t} &= \frac{V_3}{1 + (\frac{P}{K_{i3}})^{n_{i3}} + (\frac{K_{a3}}{M_2})^{n_{a3}}} - k_3 \cdot G_3 \\
    \frac{d E_1}{d t} &= \frac{V_4 \cdot G_1}{K_4 + G_1} + k_4 \cdot E_1 \\
    \frac{d E_2}{d t} &= \frac{V_5 \cdot G_2}{K_5 + G_2} + k_5 \cdot E_2 \\
    \frac{d E_3}{d t} &= \frac{V_6 \cdot G_3}{K_6 + G_3} + k_6 \cdot E_3 \\
    \frac{d M_1}{d t} &= \frac{k_{cat1} \cdot E_1 \cdot \frac{1}{K_{m1}}\cdot (S-M_1)}{1 + \frac{S}{K_{m1}} + \frac{M_1}{K_{m2}}} - \frac{k_{cat2} \cdot E_2 \cdot \frac{1}{K_{m3}}\cdot (M_1-M_2)}{1 + \frac{M_1}{K_{m3}} + \frac{M_2}{K_{m4}}}\\
    \frac{d M_2}{d t} &= \frac{k_{cat2} \cdot E_2 \cdot \frac{1}{K_{m3}}\cdot (M_1-M_2)}{1 + \frac{M_1}{K_{m3}} + \frac{M_2}{K_{m4}}} - \frac{k_{cat3} \cdot E_3 \cdot \frac{1}{K_{m5}}\cdot (M_2-P)}{1 + \frac{M_2}{K_{m5}} + \frac{P}{K_{m6}}}\\
\end{split}
\right.
\end{equation}

Specific parameter values can be found in \cite{engl2009inverse}. We take the modeling domain to be $\Omega =[0,1]^8.$ Our prior model is obtained by linearizing the mRNA ($E_1, E_2,E_3$) and metabolite ($M_1, M_2$) dynamics. We also suppose the carrying capacity parameters are incorrectly specified as $n_{i1} = n_{i2} = n_{i3} = n_{a1} = n_{a2} = n_{a3} = 1$ while for the true model they are $n_{i1} = n_{i2} = n_{i3} = n_{a1} = n_{a2} = n_{a3} = 2$. Our prior model is the following, 

\begin{equation}
\left\{
\begin{split}
    \frac{d G_1}{d t} &= \frac{V_1}{1 + (\frac{P}{K_{i1}})^{n_{i1}} + (\frac{K_{a1}}{S})^{n_{a1}}} - k_1 \cdot G_1 \\
    \frac{d G_2}{d t} &= \frac{V_2}{1 + (\frac{P}{K_{i2}})^{n_{i2}} + (\frac{K_{a2}}{M_1})^{n_{a2}}} - k_2 \cdot G_2 \\
    \frac{d G_3}{d t} &= \frac{V_3}{1 + (\frac{P}{K_{i3}})^{n_{i3}} + (\frac{K_{a3}}{M_2})^{n_{a3}}} - k_3 \cdot G_3 \\
    \frac{d E_1}{d t} &= {V_4 \cdot G_1} + k_4 \cdot E_1 \\
    \frac{d E_2}{d t} &= {V_5 \cdot G_2} + k_5 \cdot E_2 \\
    \frac{d E_3}{d t} &= {V_6 \cdot G_3} + k_6 \cdot E_3 \\
    \frac{d M_1}{d t} &= {k_{cat1} \cdot E_1 \cdot \frac{1}{K_{m1}}\cdot (S-M_1)} - {k_{cat2} \cdot E_2 \cdot \frac{1}{K_{m3}}\cdot (M_1-M_2)}\\
    \frac{d M_2}{d t} &= {k_{cat2} \cdot E_2 \cdot \frac{1}{K_{m3}}\cdot (M_1-M_2)}  - {k_{cat3} \cdot E_3 \cdot \frac{1}{K_{m5}}\cdot (M_2-P)}.\\
\end{split}\\
\right.
\label{metabolic_prior}
\end{equation}

The prior model \eqref{metabolic_prior} is used to generate $75,000$ data pairs over a time lag of $\Delta t = 0.05$ and data pairs are randomly distributed within the time frame $T=12.5$. The DNN consists of 80 neurons per layer. After training the DNN prior model for $20,000$ epochs, we conduct model correction using 750 high-fidelity data pairs generated by the true model.

\begin{figure}
    \centering
{\includegraphics[width=0.40\textwidth]{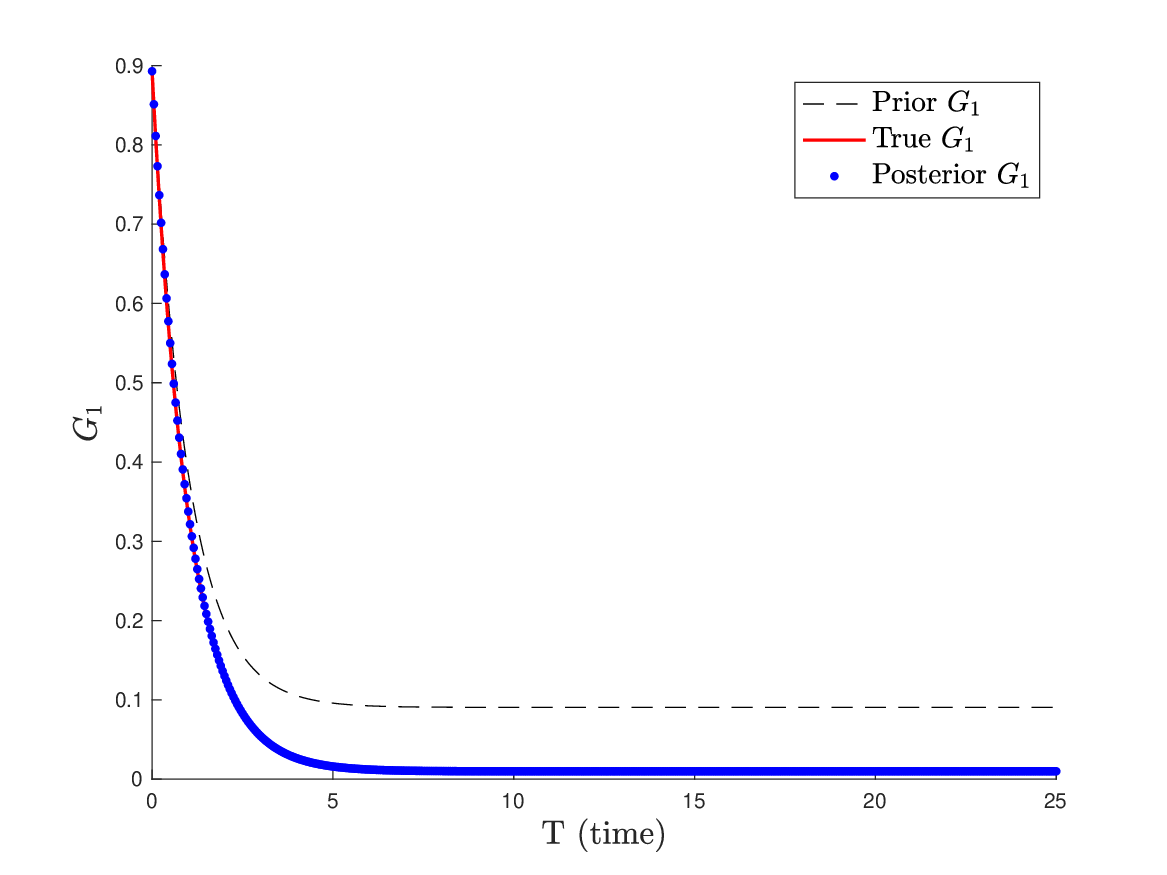}}
\hfil
{\includegraphics[width=0.40\textwidth]{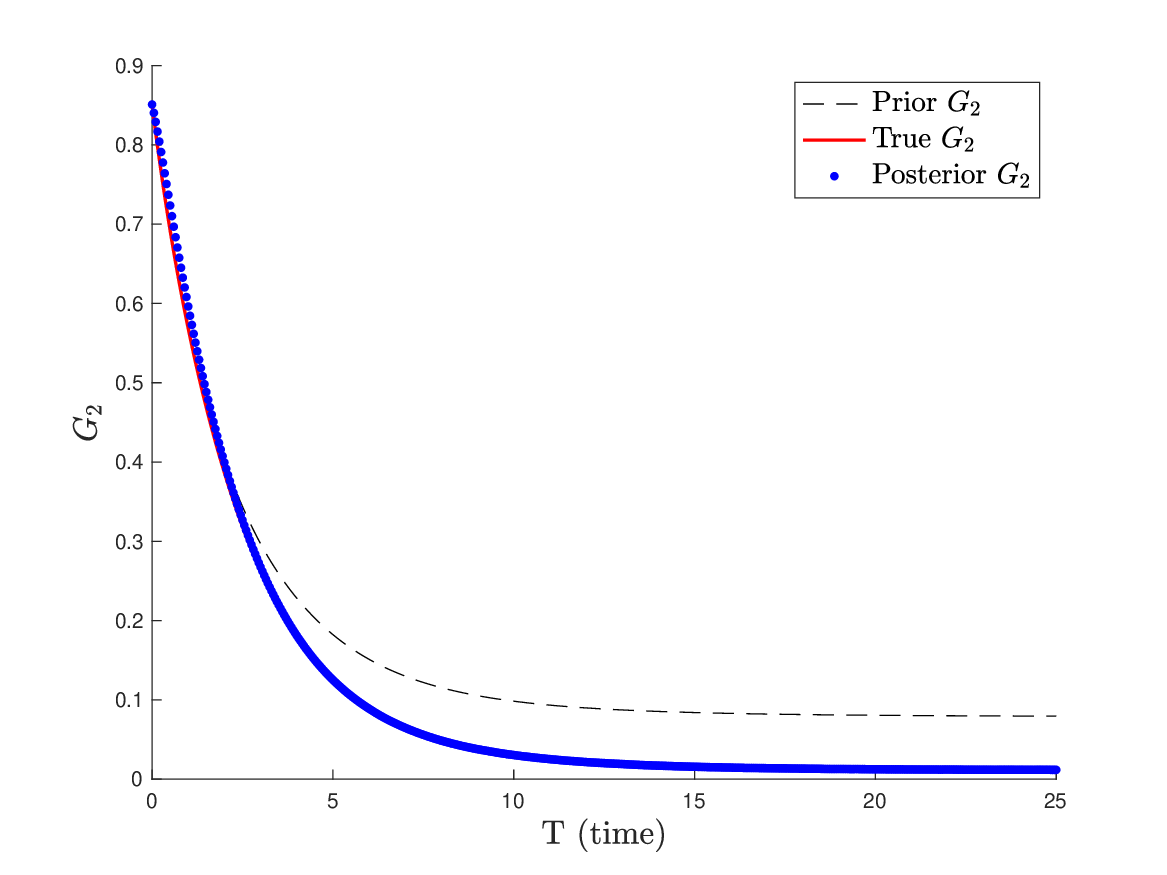}}

\medskip
{\includegraphics[width=0.40\textwidth]{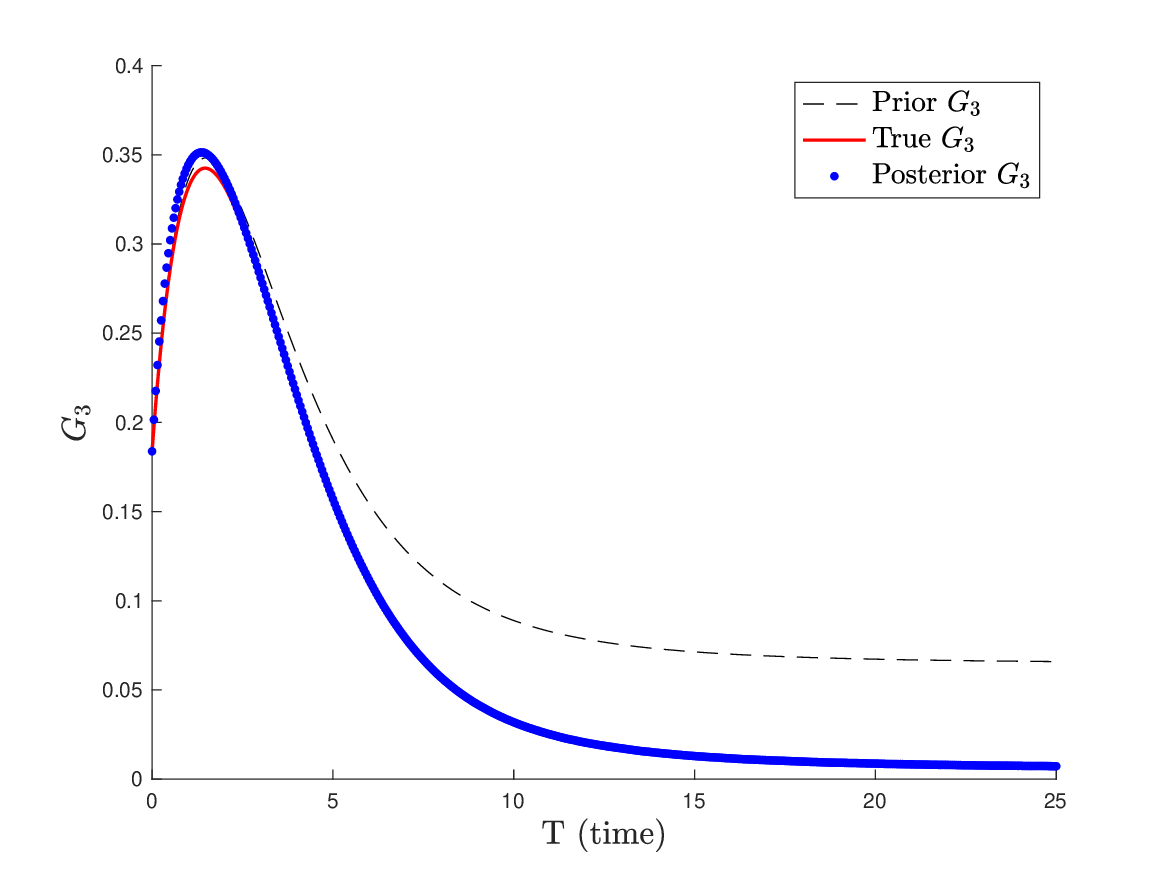}}
\hfil
{\includegraphics[width=0.40\textwidth]{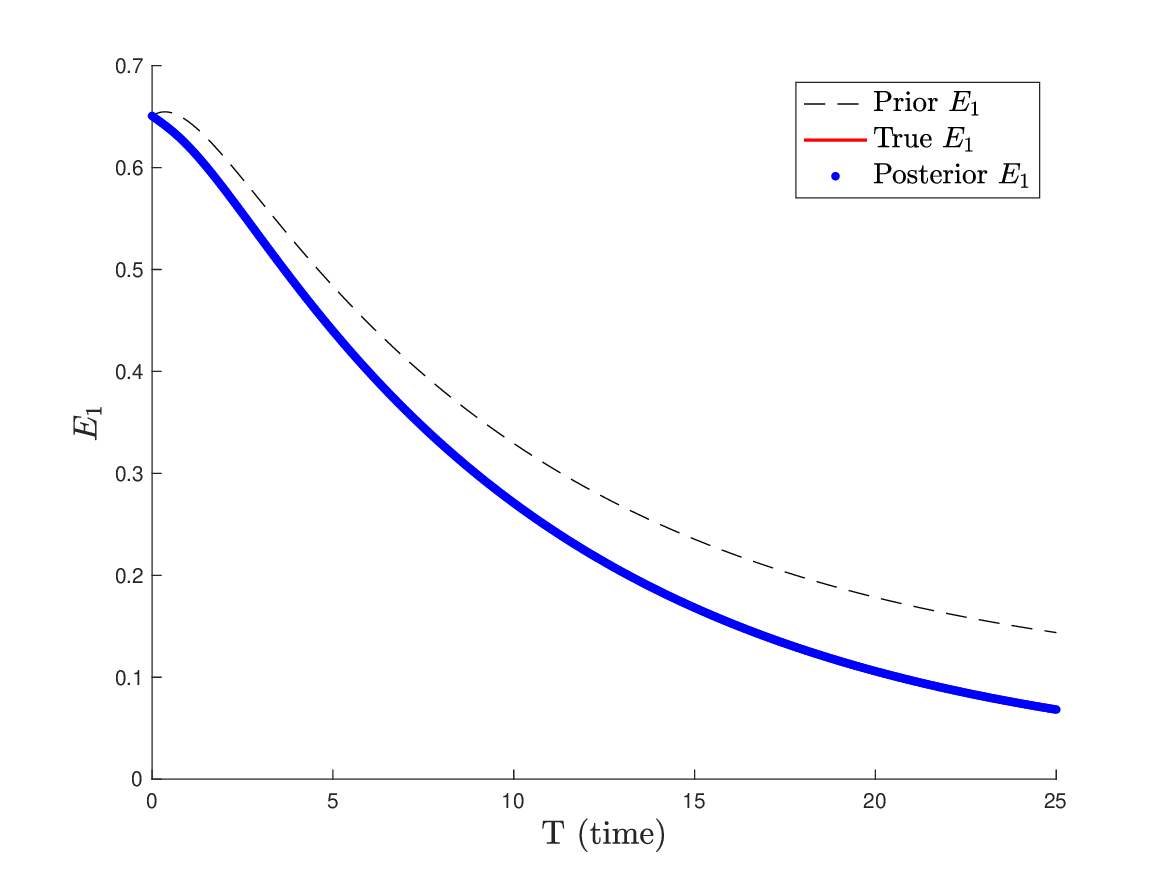}}

\medskip
{\includegraphics[width=0.40\textwidth]{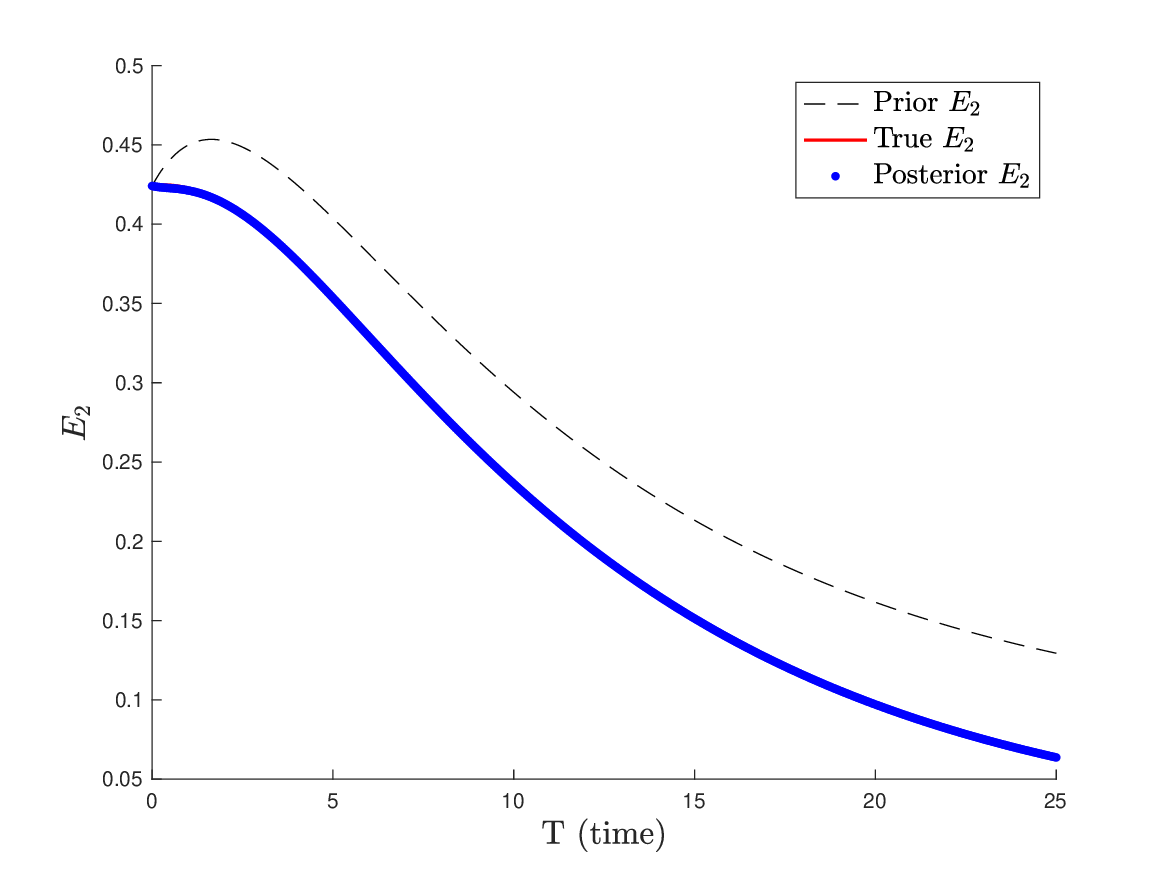}}
\hfil
{\includegraphics[width=0.40\textwidth]{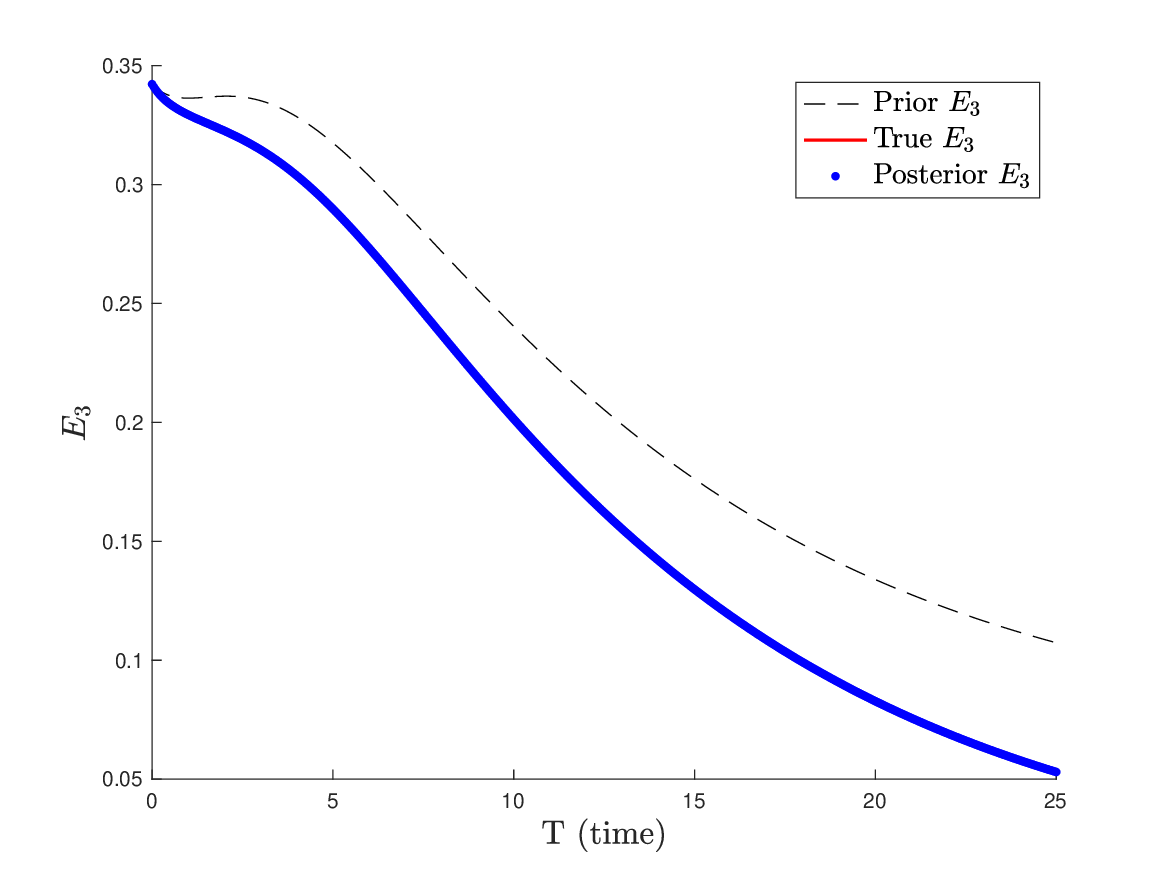}}

\medskip
{\includegraphics[width=0.40\textwidth]{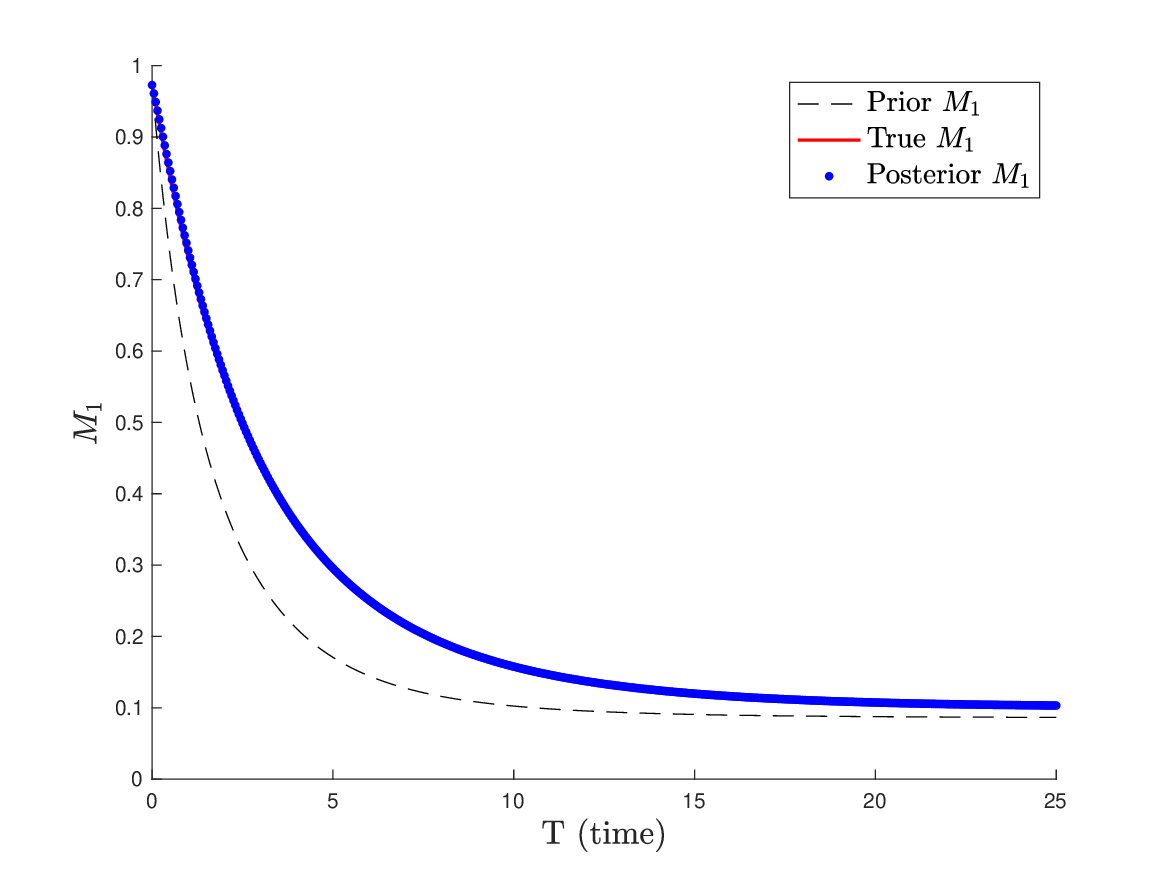}}
\hfil
{\includegraphics[width=0.40\textwidth]{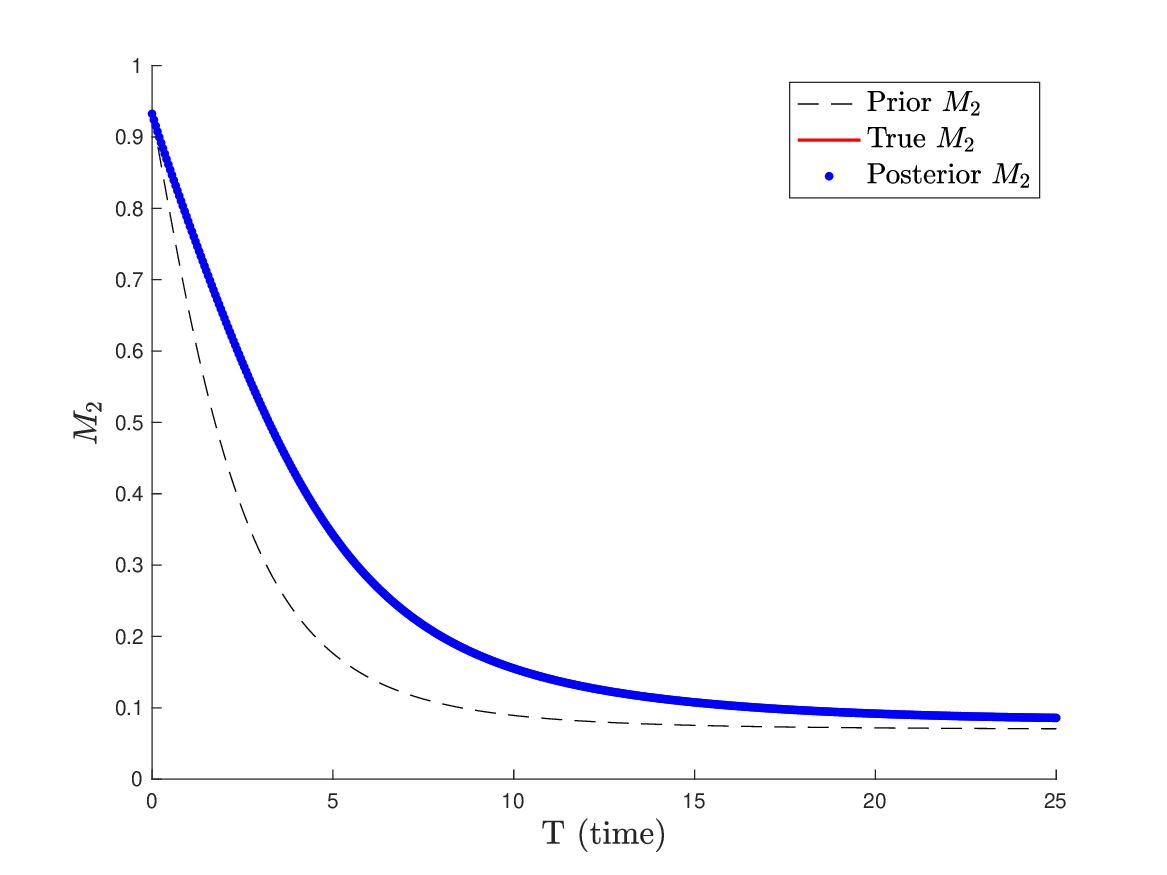}}
\caption{Metabolic pathway: system prediction up to $T=25$. Randomly selected initial condition: $G_1 = 0.893, G_2 = 0.851 , G_3 = 0.184, E_1 = 0.651, E_2 = 0.424, E_3 =0.342, M_1 = 0.973, M_2 = 0.932$.}
    \label{metabolic_example}
    \end{figure}

\begin{figure} 
\centering
\includegraphics[width = \textwidth]{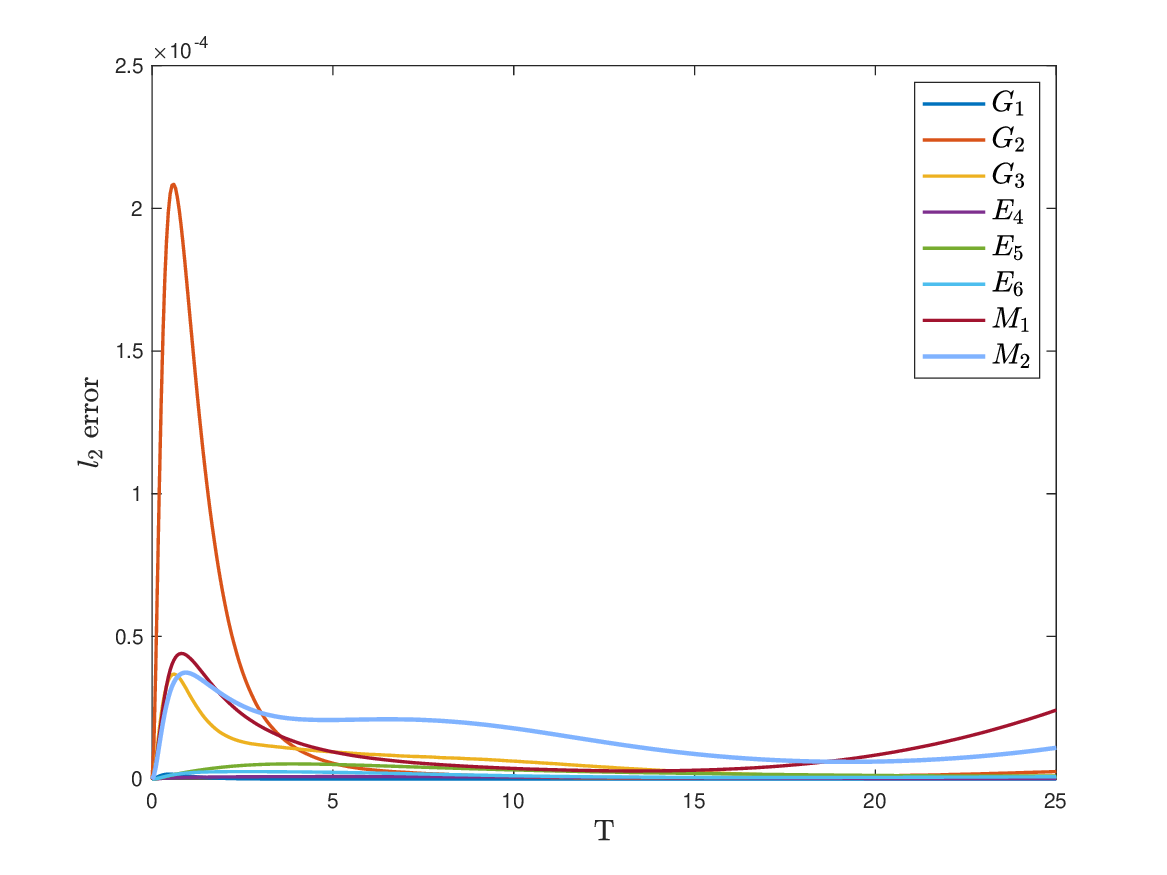}
    \caption{Metabolic pathway: average $l_2$ error over 100 test trajectories.}
    \label{metabolic_error}
\end{figure}

Figure \ref{metabolic_example} depicts predictions made up to $T = 20$ for a randomly selected initial condition. In Figure \ref{metabolic_error} we plot the average prediction error for the $8$ state variables over $100$ test trajectories. For this complex biological system example, the posterior DNN provides good accuracy and visible improvement over the prior model prediction.

\subsection{Coarse data observations}\label{subsec:sec4.2} 

We now present four examples demonstrating the method outlined in section 2.3.3 for handling scarce as well as coarse high-fidelity data. The DNNs have 5 hidden layers while the number of neurons in the first three layers contain $50$ neurons and the final two layers contain $20$ or $50$ neurons. 
The DNN prior model is trained using data generated by the low-fidelity model on the desired time scale $\delta$ for $10,000$ epochs, using a batch size of $100$ and early-stopping with a patience of $1,000$ epochs to avoid over-fitting. Transfer learning is then conducted on the final two layers of the DNN model, $\mathbf{W}_{[M-1:M]}$, using \eqref{TL-loss-recurrent-loss}.

\subsubsection{Damped Pendulum} We revisit example 4.1.2, where the true model is a damped pendulum system, 
\begin{equation}
\left\{
\begin{split}
    \dot{x}_1 & = x_2\\
    \dot{x}_2 & = -\alpha x_2 - \beta \sin(x_1).  
\label{damped_coarse}
    \end{split}
    \right.
    \end{equation}
and the prior model is the linear harmonic oscillator,

\begin{equation}
\left\{
\begin{split}
    \dot{x}_1 & = x_2\\
    \dot{x}_2 & = -\beta x_1. 
\end{split}
\right.
\label{damped_coarse_prior}
\end{equation}
    Data pairs are sampled over the modeling domain $\Omega =[-2\pi,2\pi] \times [-\pi,\pi]$ and are generated by randomly selecting an initial condition from $\Omega$ and evolving the state variables over the desired time step $\delta = 0.2$. The DNN is taken to be 5 hidden layers; the first 3 layers contain 50 neurons and then final two layers contain 20 neurons. The DNN is first trained with 50,000 data pairs generated by the prior model \eqref{damped_coarse_prior} for 10,000 epochs. 
    
We then use 500 high-fidelity data pairs generated by the true system \eqref{damped_coarse} for model correction; each data pair is separated by a coarse time-lag $\Delta^{(j)}$, randomly sampled from the set $\Delta^{(j)} \in \{1, 1.2,\dots, 9.8, 10\}$. 

\begin{figure}
\centering

  \includegraphics[width=0.49\linewidth]{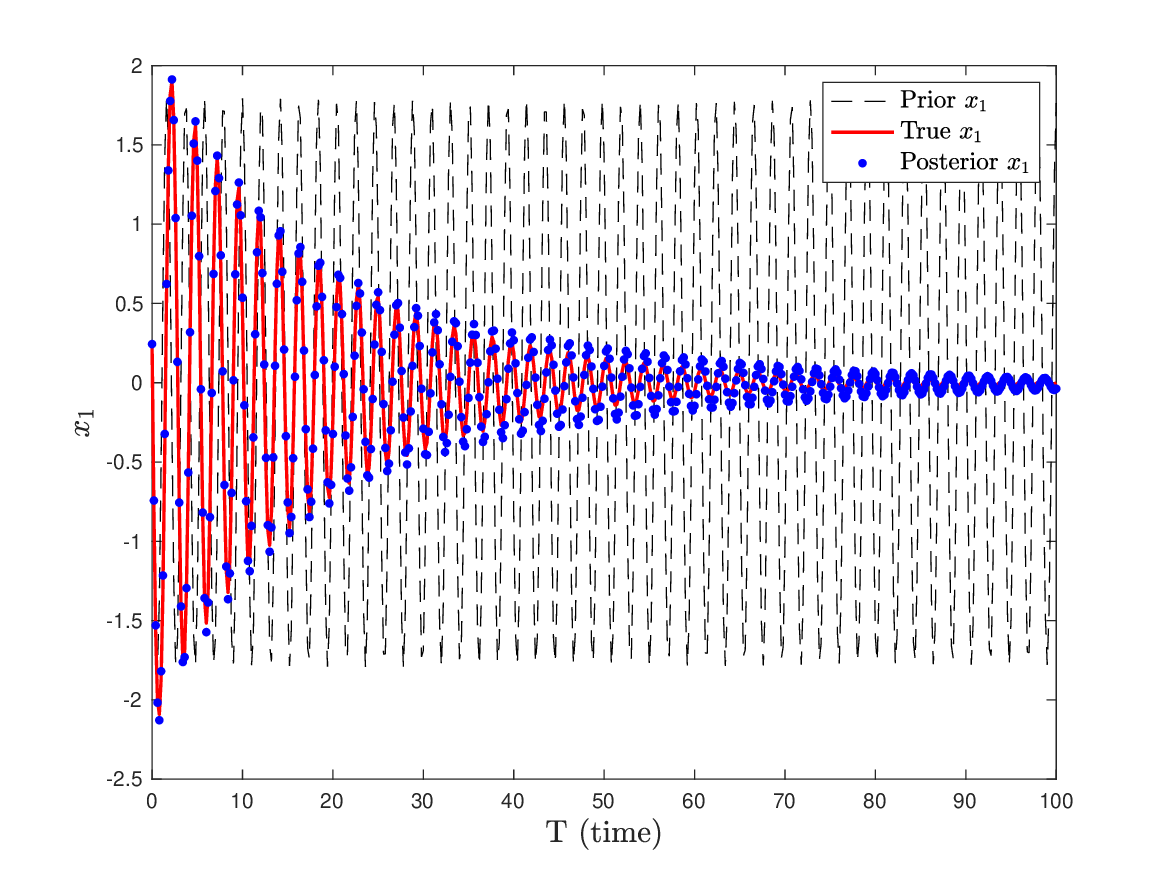}
  \includegraphics[width=0.49\linewidth]{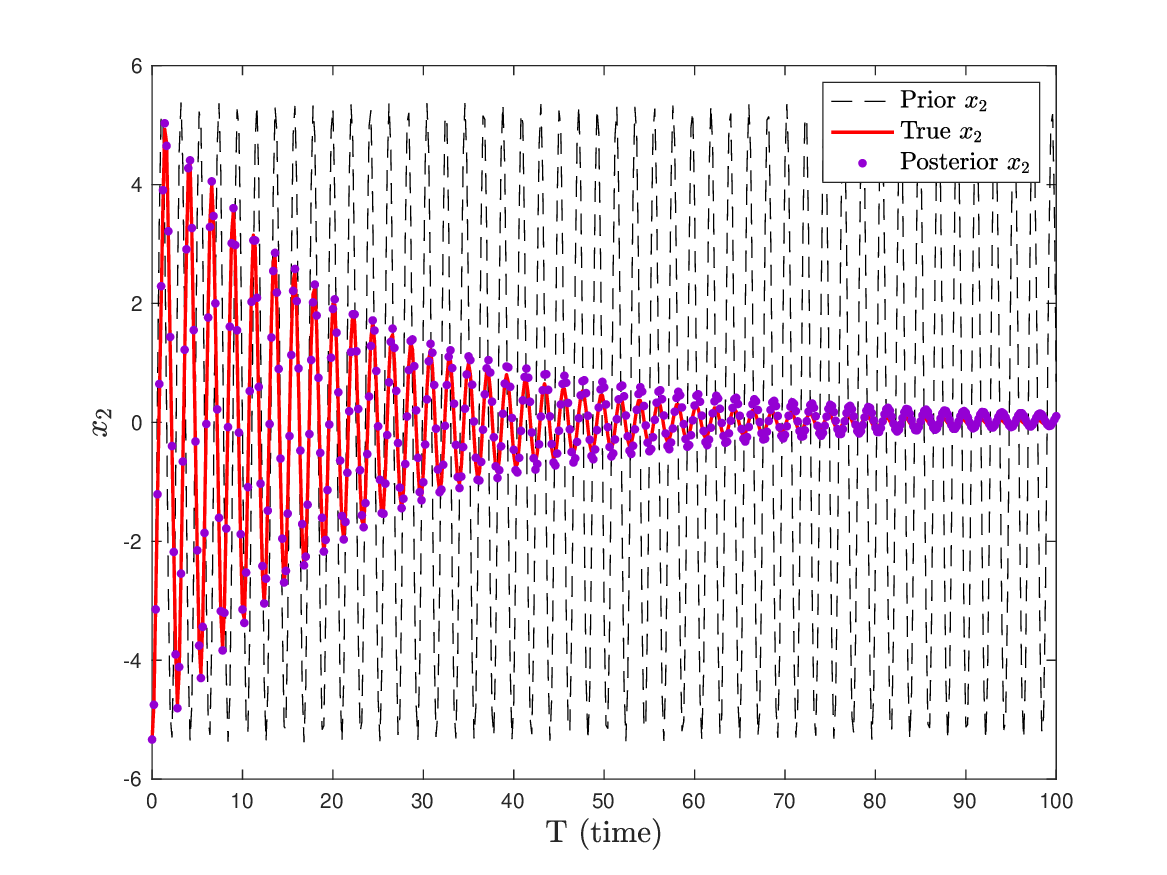}

\caption{Damped pendulum: system prediction up to $T=100$. Left: $x_1$; Right: $x_2$.}
\label{fig:damped_coarse}
\end{figure}

\begin{figure}
    \centering \includegraphics[width=0.75\linewidth]{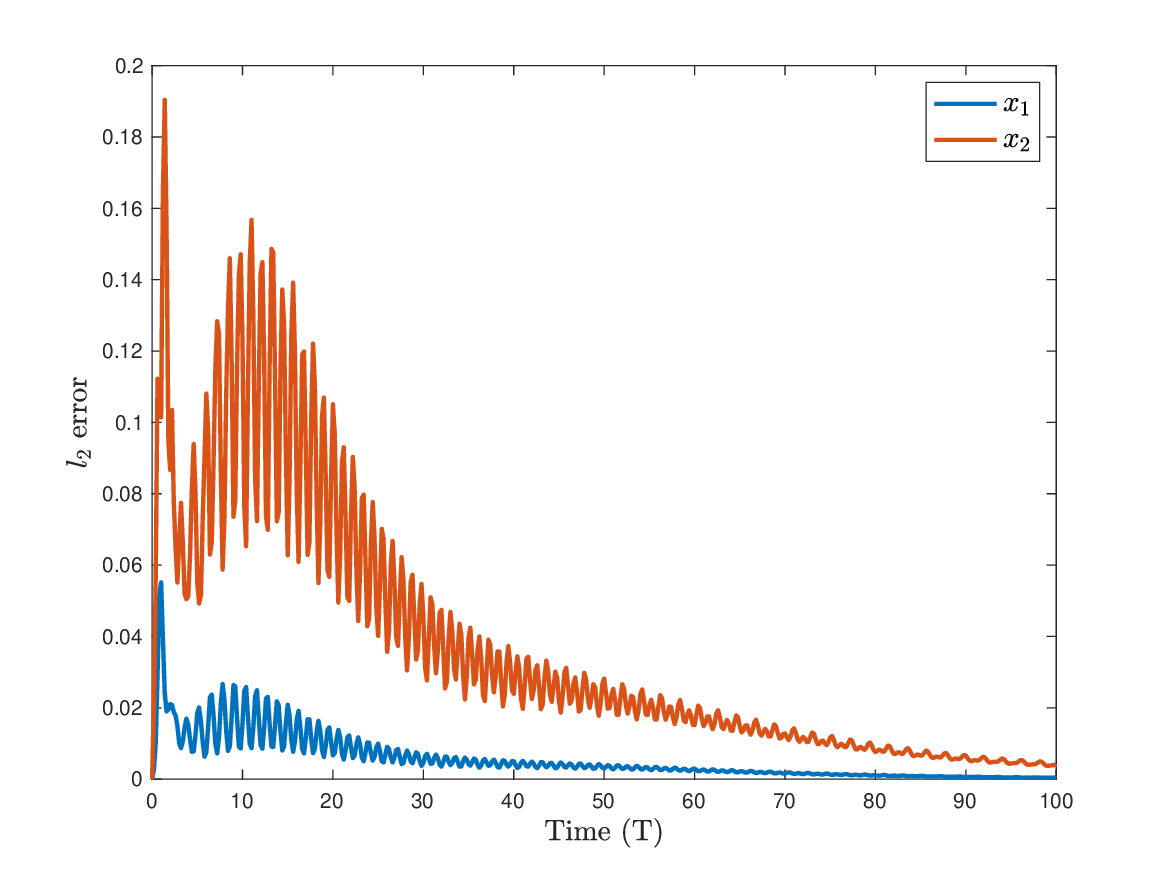}
    \caption{Damped pendulum: average $l_2$ error over 100 test trajectories.}
    \label{fig:damped_coarse_error}
\end{figure}

 Figure \ref{fig:damped_coarse} presents an example trajectory prediction up to $T=100$ with the randomly selected initial condition $\mathbf{x}_0 = (-0.242, -4.241)$. Predictions are conducted on the desired time scale $\delta = 0.2$.  Figure \ref{fig:damped_coarse_error} shows the average error over 100 test trajectories and demonstrates good accuracy, despite using only coarse and scarce high-fidelity data for model correction.

\subsubsection{Van der Pol oscillator} 

Our next example is the Van der Pol oscillator. In this system, the strength of the non-linearity is controlled by the scalar factor $\mu$, 
\begin{equation}
\left\{
\begin{split}
         \dot{x}_1 & = x_2\\
        \dot{x}_2 & = \mu(1-x_1^2)x_2 - x_1.  
\end{split}
\right.
\end{equation}
For the true model we set $\mu = 1$, while the prior model assumes $\mu = 0.5$. 

We consider the modeling domain $\Omega =[-2,2] \times [-1.5,1.5]$. In this example, 50,000 data pairs are generated by the prior model over the time scale $\delta = 0.2$, and are randomly distributed within the time frame $T = 20$. The final 2 layers of the DNN prior model contains 20 neurons. 

\begin{figure}
\centering

    \includegraphics[width=0.49\linewidth]{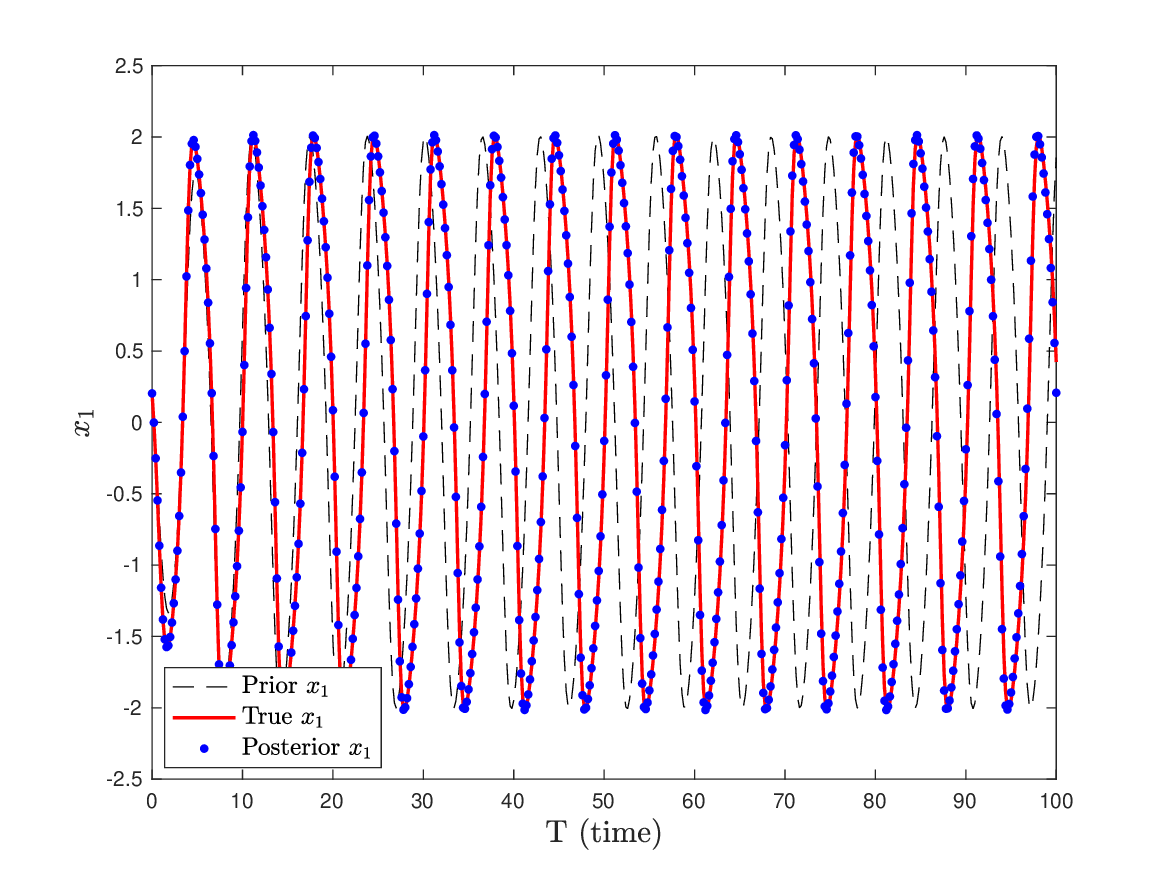}
  \includegraphics[width=0.49\linewidth]{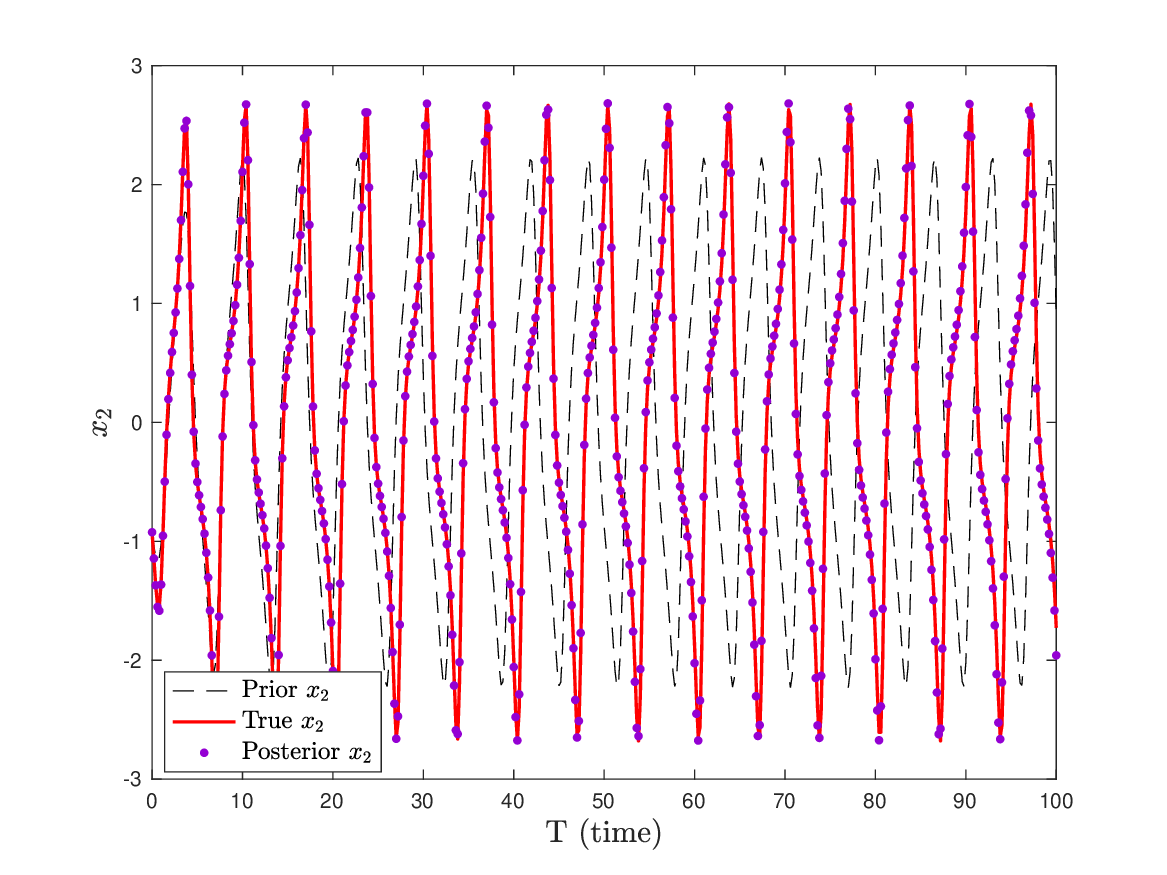}

\caption{Van der Pol oscillator: system prediction up to $T=100$. Left: $x_1$; Right: $x_2$.}
\label{van_der_pol}
\end{figure}

\begin{figure}
    \centering \includegraphics[width=0.75\linewidth]{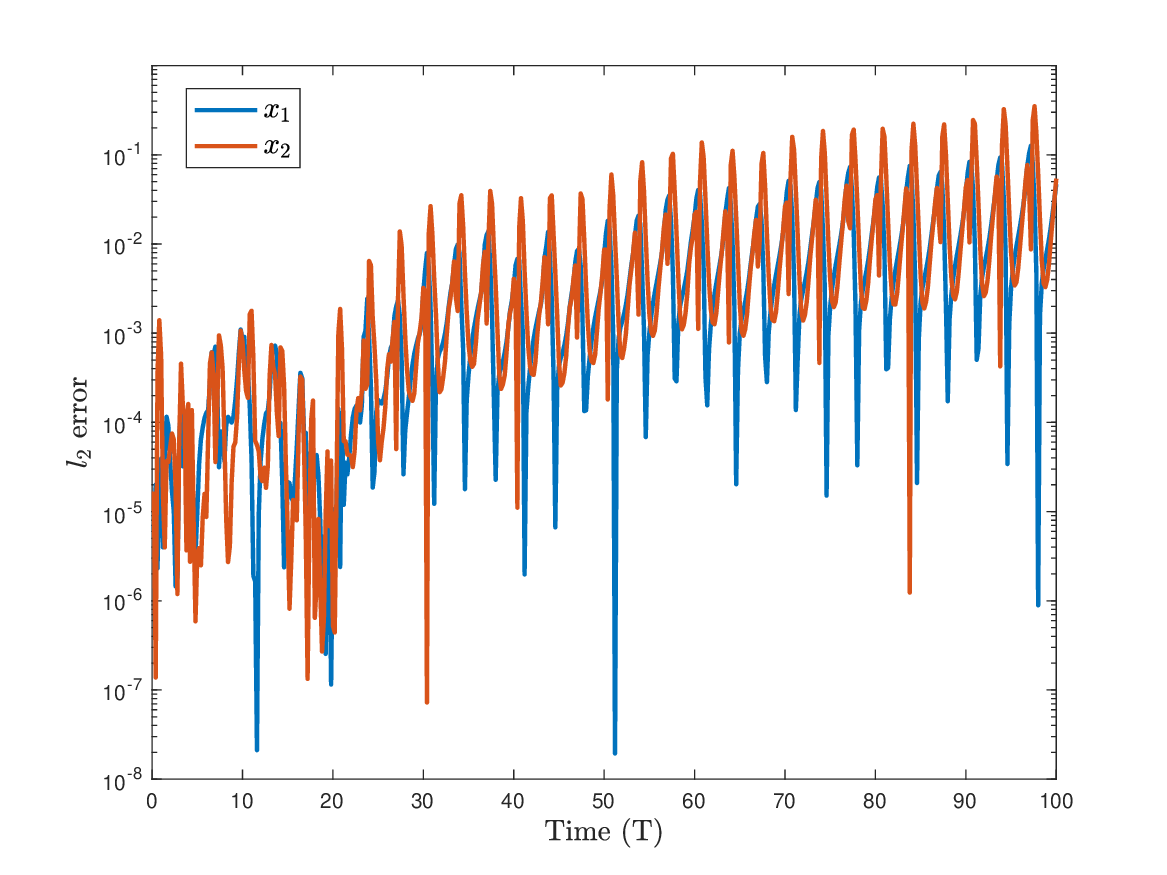}
    \caption{Van der Pol oscillator: average $l_2$ error over 100 test trajectories.}
    \label{van_pol_error}
\end{figure}

Our high-fidelity data set consists of $500$ data pairs, where each high-fidelity data pair is separated by a coarse time step randomly selected from the set $\Delta^{(j)}\in \{1, 1.2,\dots, 9.8, 10\}$. After transfer learning, DNN predictions are made up to $T = 100$, and results for a randomly selected initial condition $\mathbf{x}_0 = (0.203,-0.924)$ are pictured in Figure \ref{van_der_pol}. Figure \ref{van_pol_error} depicts the average error over 100 trajectories and we note again the ability of the network to correct the prior model for relatively long term predictions. 

\subsubsection{Differential-algebraic system} Our next example is the four-dimensional system of nonlinear differential-algebraic equations modeling an electric network, 
\begin{equation}
\left\{
\begin{split}
   \dot{u_1} &= v_2/c \\
    \dot{u_2} &= u_1/L \\ 
    0 &= v_1 - (G_0 - G_{\inf}) U_0 {\tanh} {(u_1) -G_{\inf} u_1} \\ 
    0 &= v_2 +u_2 +v_1 .
\end{split}
\right.
\label{DAE_system}
\end{equation}

The prior model is obtained by truncating the Taylor series of the hyperbolic tangent function for a cubic approximation in the first algebraic condition,   
\begin{equation}
\left\{
\begin{split}
   		\dot{u_1} &= v_2/c \\
		\dot{u_2} &= u_1/L \\ 
            0 &= v_1 - (G_0 - G_{\inf}) U_0 (u_1 - u_1^3) - G_{\inf} u_1 \\ 
            0 &= v_2 +u_2 +v_1 .
\end{split}
\right.
\label{prior_DAE_system}
\end{equation}

The node voltage is represented by $u_1$ while $u_2$, $v_1$ and $v_2$ are branch currents. The parameter values are defined as $C = 10^{-9}, L = 10^{-6}, U_0 = 1, G_0 = -0.1$ and $G_{\infty} = 0.25$.  

\begin{figure}
    \centering
{\includegraphics[width=0.49\textwidth]{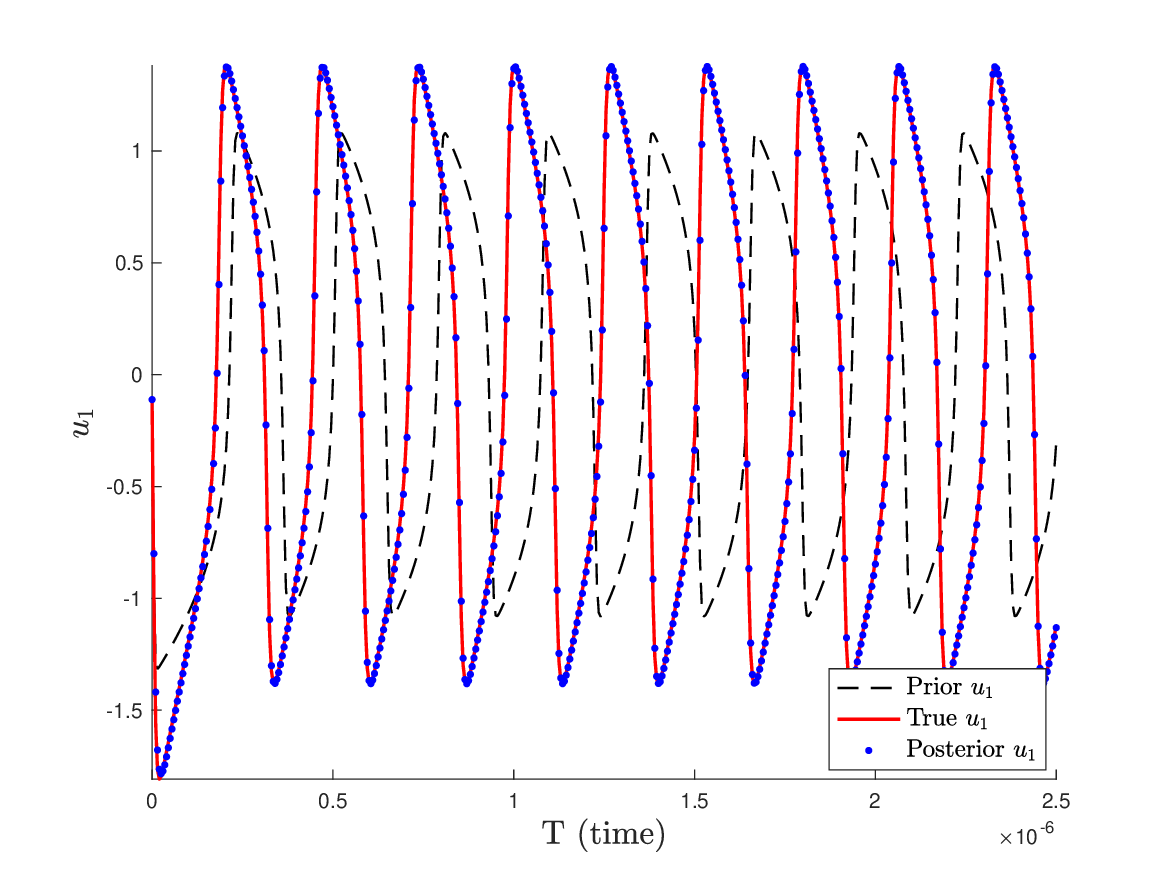}\label{u1_fig}}
\hfil
{\includegraphics[width=0.49\textwidth]{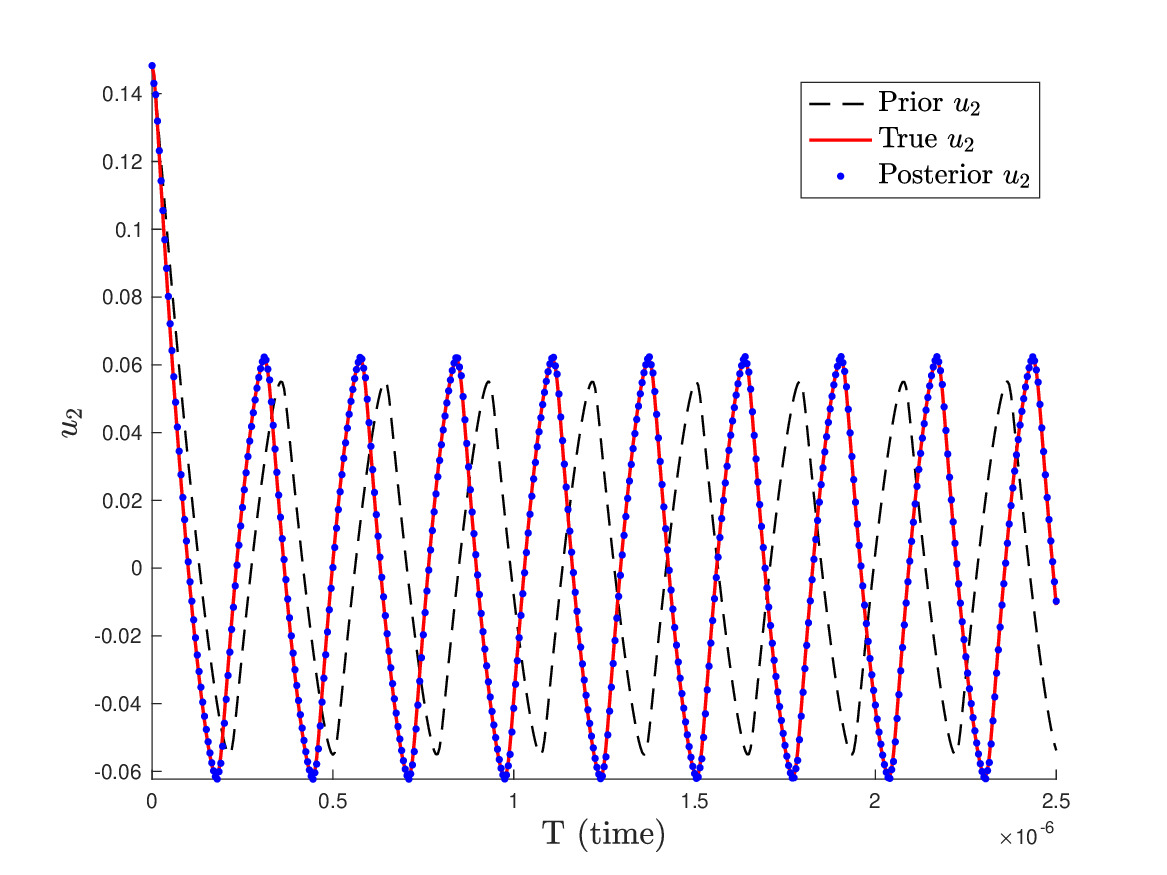}\label{u2_fig}}

\medskip
{\includegraphics[width=0.49\textwidth]{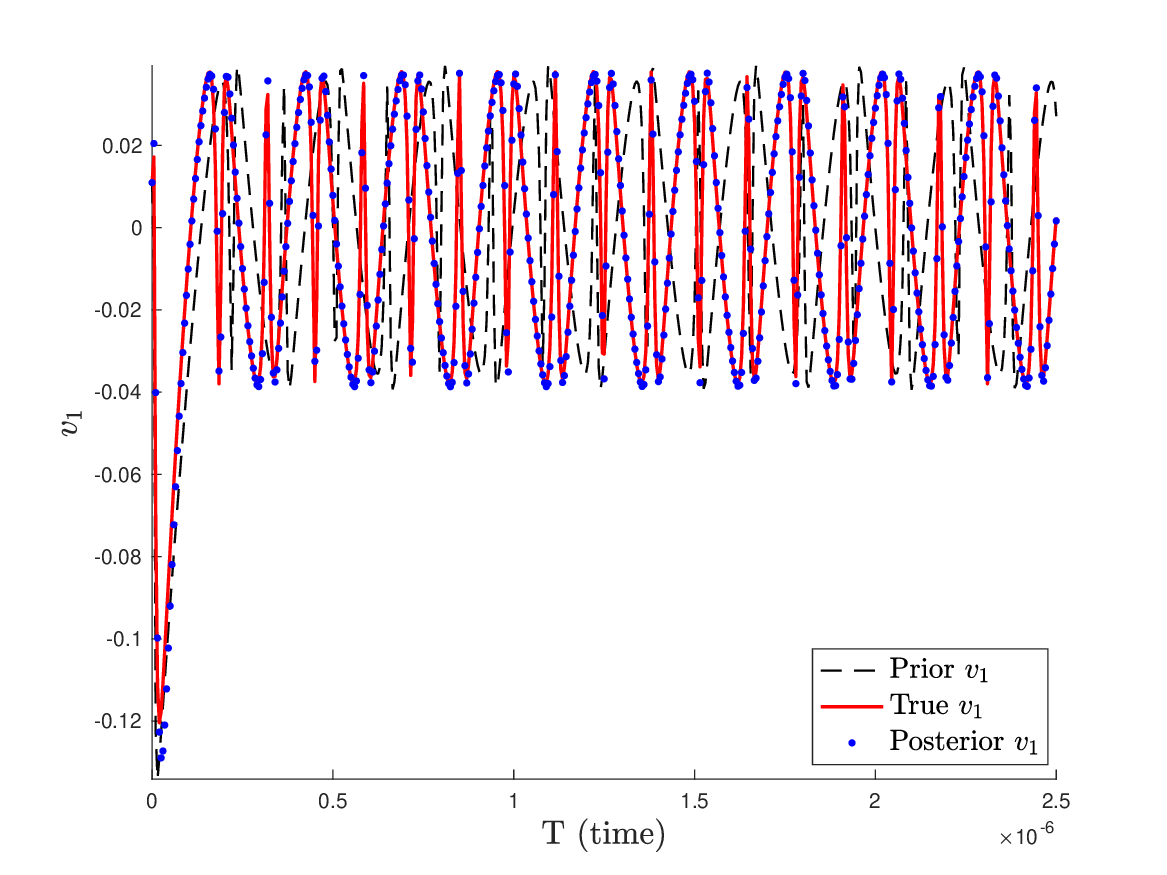}\label{u3_fig}}
\hfil
{\includegraphics[width=0.49\textwidth]{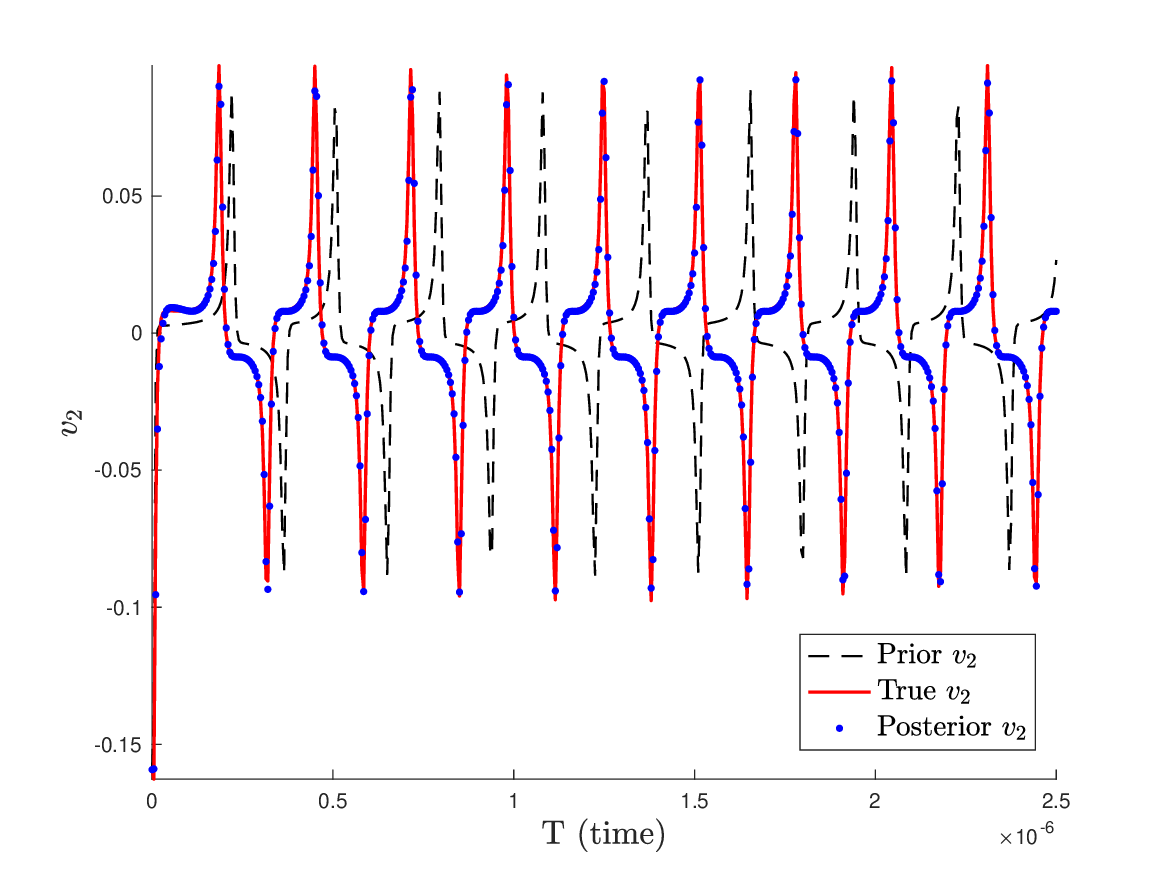}\label{u4_fig}}

\caption{Differential algebraic system: system prediction up to $T = 2.5 \times 10^{-6}$. Top left: $u_1$. Top right: $u_2$. Bottom left: $v_1$. Bottom right: $v_2$. }

    \label{DAE_example_traj}
    \end{figure}

\begin{figure}
\centering
\includegraphics[width = \textwidth]{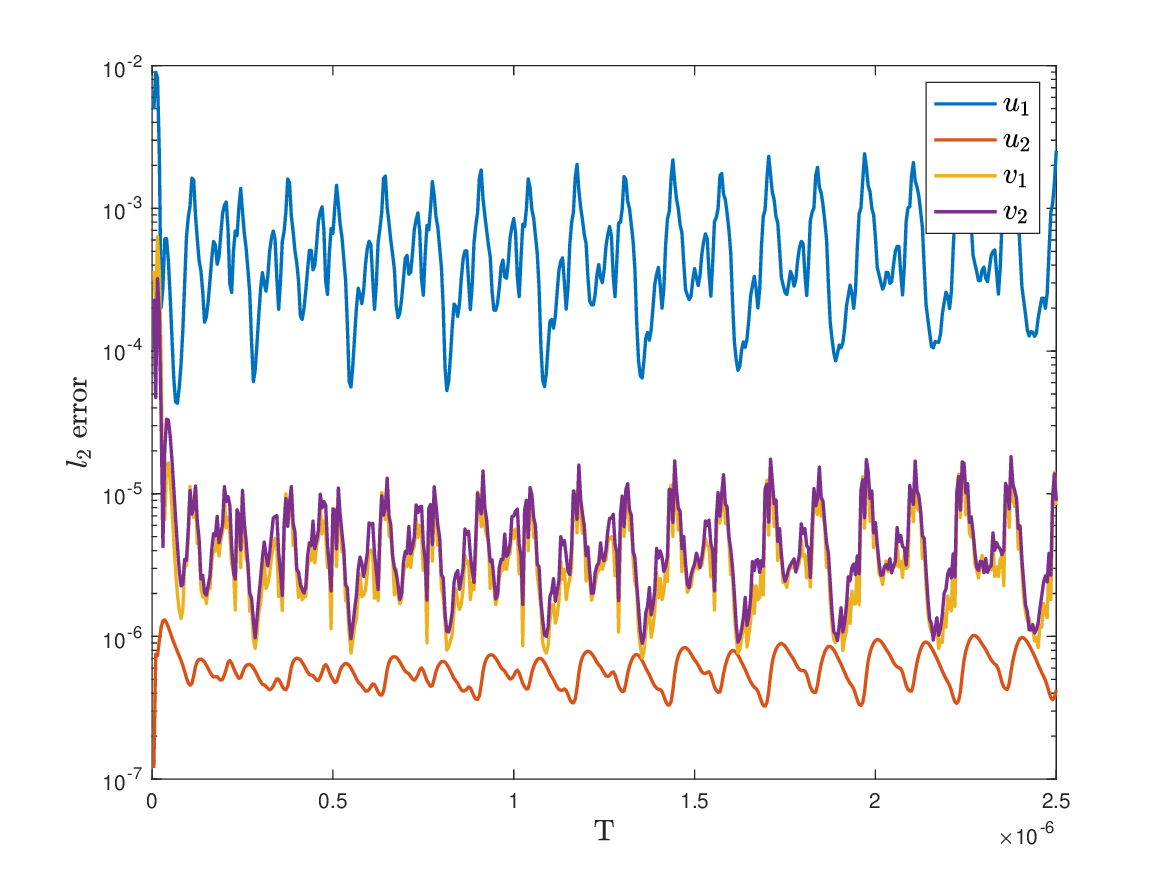}
    \caption{DAE system: average $l_2$ error over 100 test trajectories.}
    \label{DAE_error}
\end{figure}

 The modeling domain is $\Omega =[-2,2] \times [-0.2,0.2]$.  We generate $60,000$ data pairs using the prior model \eqref{prior_DAE_system} over the time step $\delta = 5\times10^{-9}$, which are sampled randomly from trajectories of length $T = 5 \times 10^{-7}$. The DNN model is 5 hidden layers with 50 nodes per layer. The high-fidelity data is $500$ data pairs generated by the true underlying system and separated by a coarse time lag randomly selected from the set $\Delta_{k^{(j)}} \in \{2.5\times 10^{-8}, 3.0\times 10^{-8},\dots,1.5\times 10^{-7}\}$. Figure \ref{DAE_example_traj} demonstrates the posterior model's ability to correct the complex dynamics on the desired time scale for the randomly selected initial condition $\mathbf{u}_0 = (-0.111, 0.148)$ up to $T = 2.5 \times 10^{-6}$ (500 total $\delta$ time step predictions). To validate the accuracy of our posterior model further, Figure \ref{DAE_error} depicts the average error for $100$ test trajectories. 

\subsubsection{Metabolic Pathway} 
We now reexamine the three-step metabolic pathway example presented in example 4.1.5. The modeling domain is again taken to be $\Omega =[0,1]^8$. The desired time scale is $\delta = 0.2$ and $60,000$ data pairs from the prior model are randomly generated from trajectories of length $T=20$. The DNN model consists of 5 hidden layers and 50 neurons per layer. We sample 500 high fidelity data pairs separated by a coarse time lag, which is randomly selected from the set $\Delta_{k^{(j)}} \in \{1, 1.2,\dots, 9.8, 10\}$.

\begin{figure}
    \centering
{\includegraphics[width=0.42\textwidth]{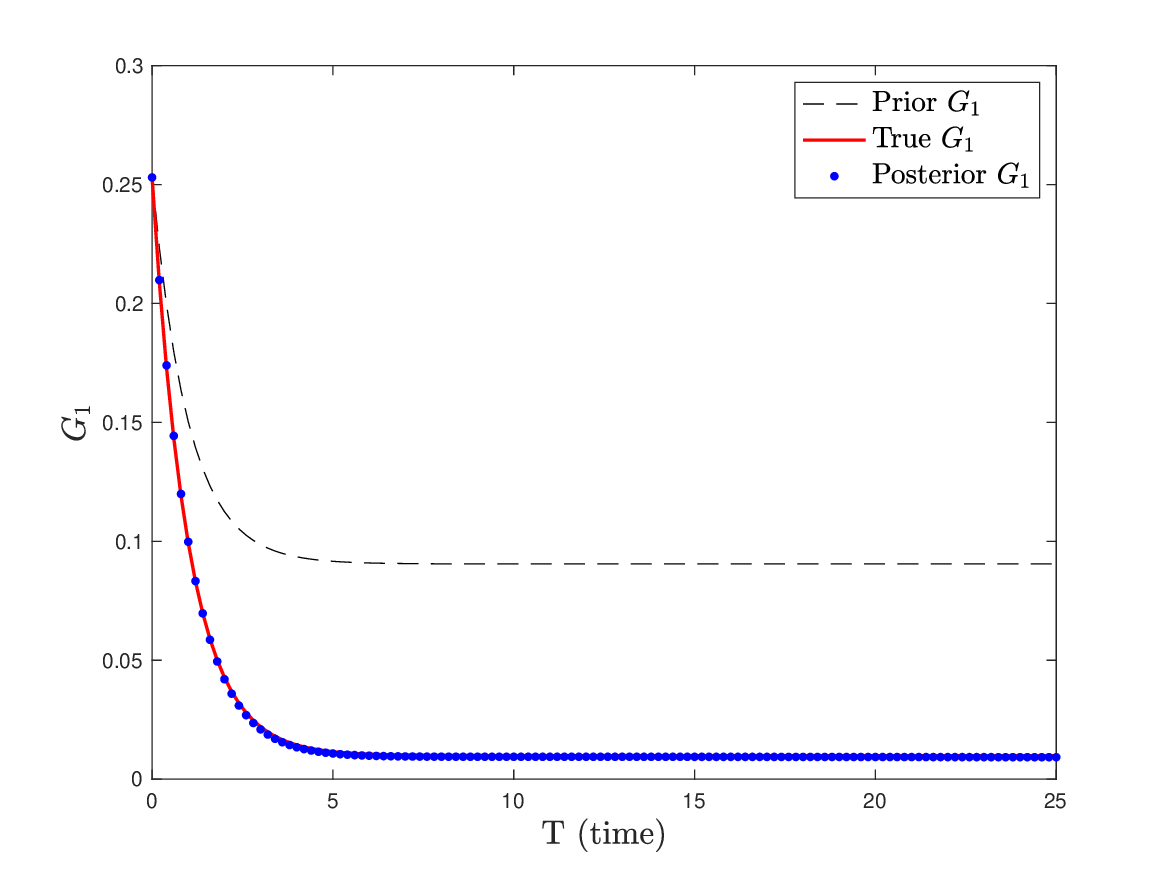}}
\hfil
{\includegraphics[width=0.42\textwidth]{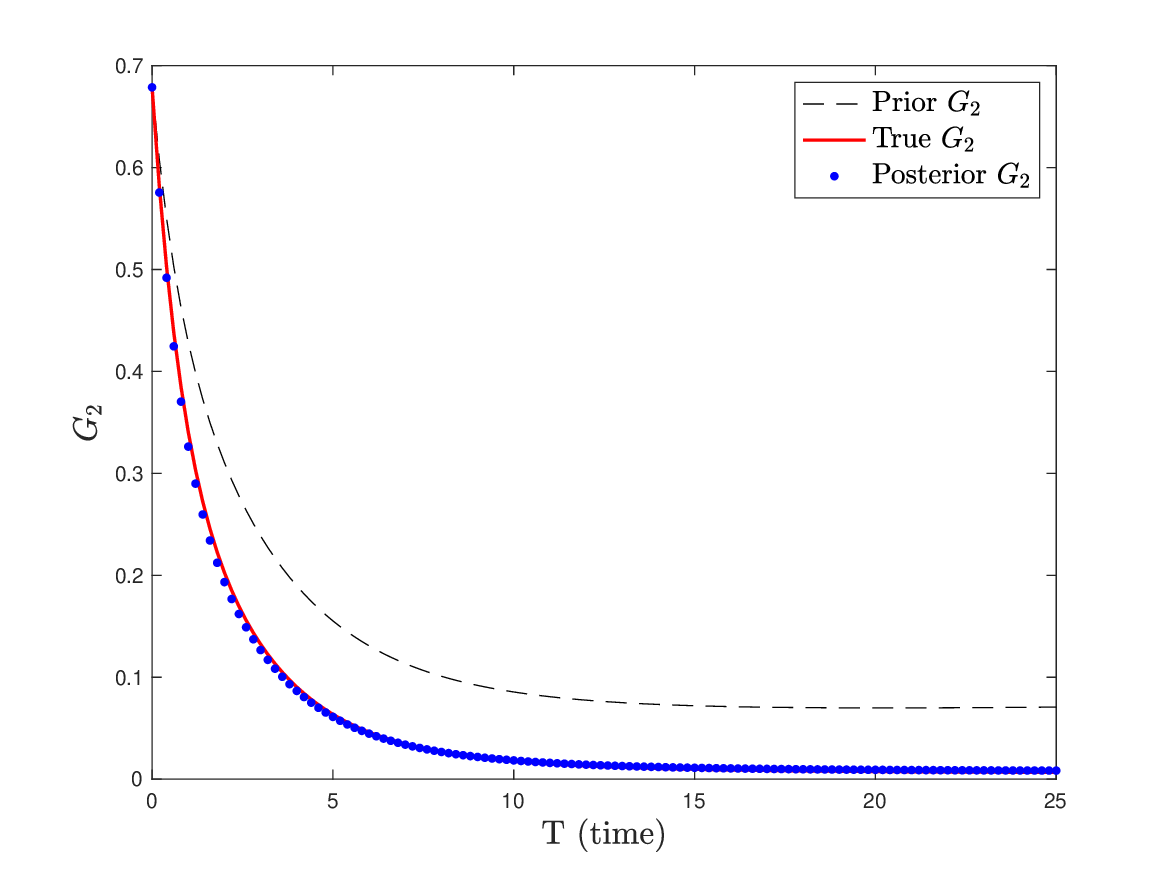}}

\medskip
{\includegraphics[width=0.42\textwidth]{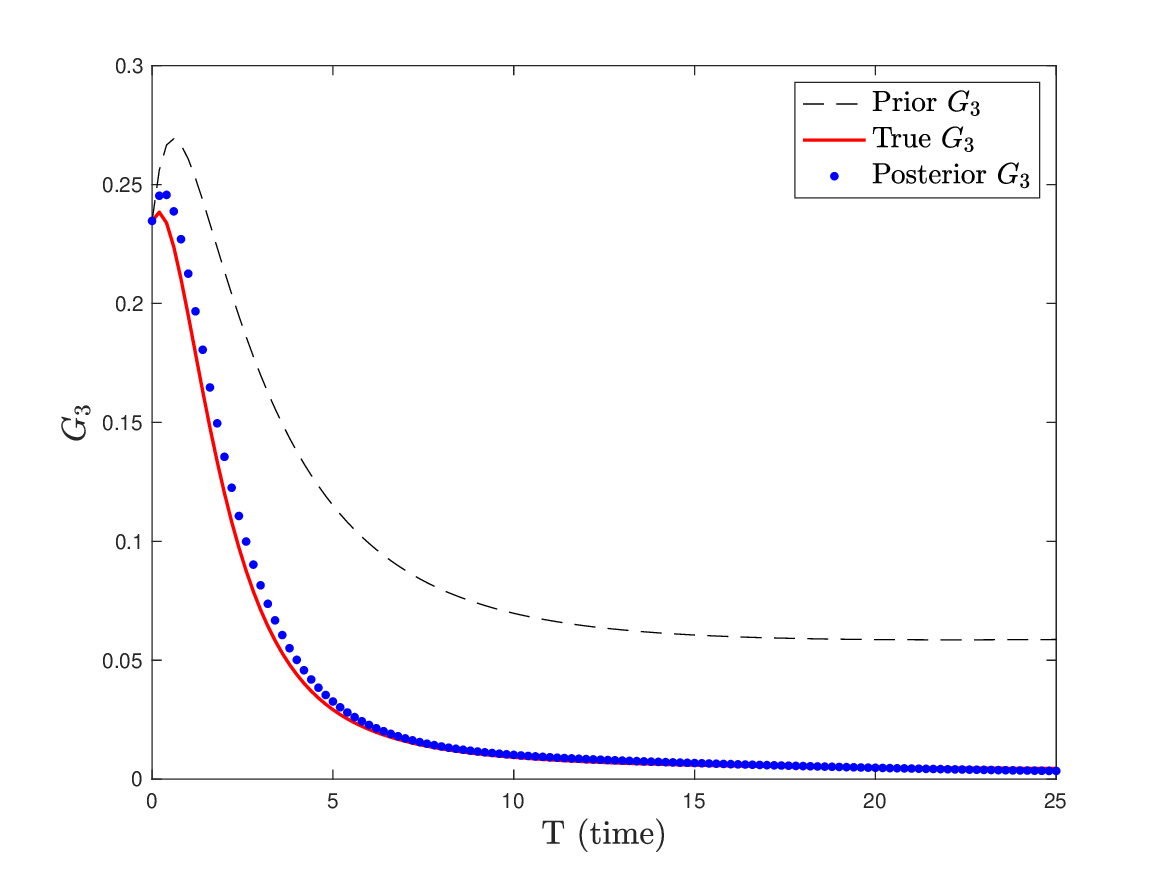}}
\hfil
{\includegraphics[width=0.42\textwidth]{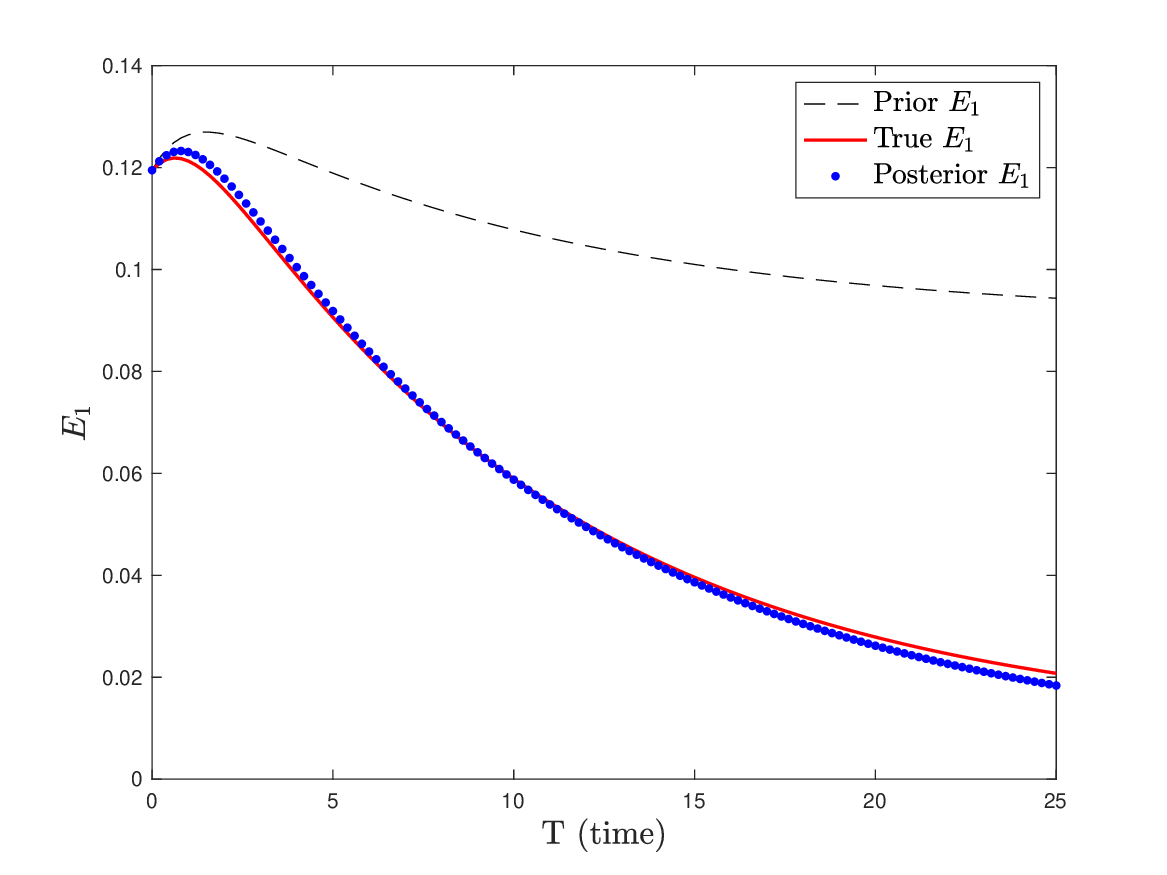}}

\medskip
{\includegraphics[width=0.42\textwidth]{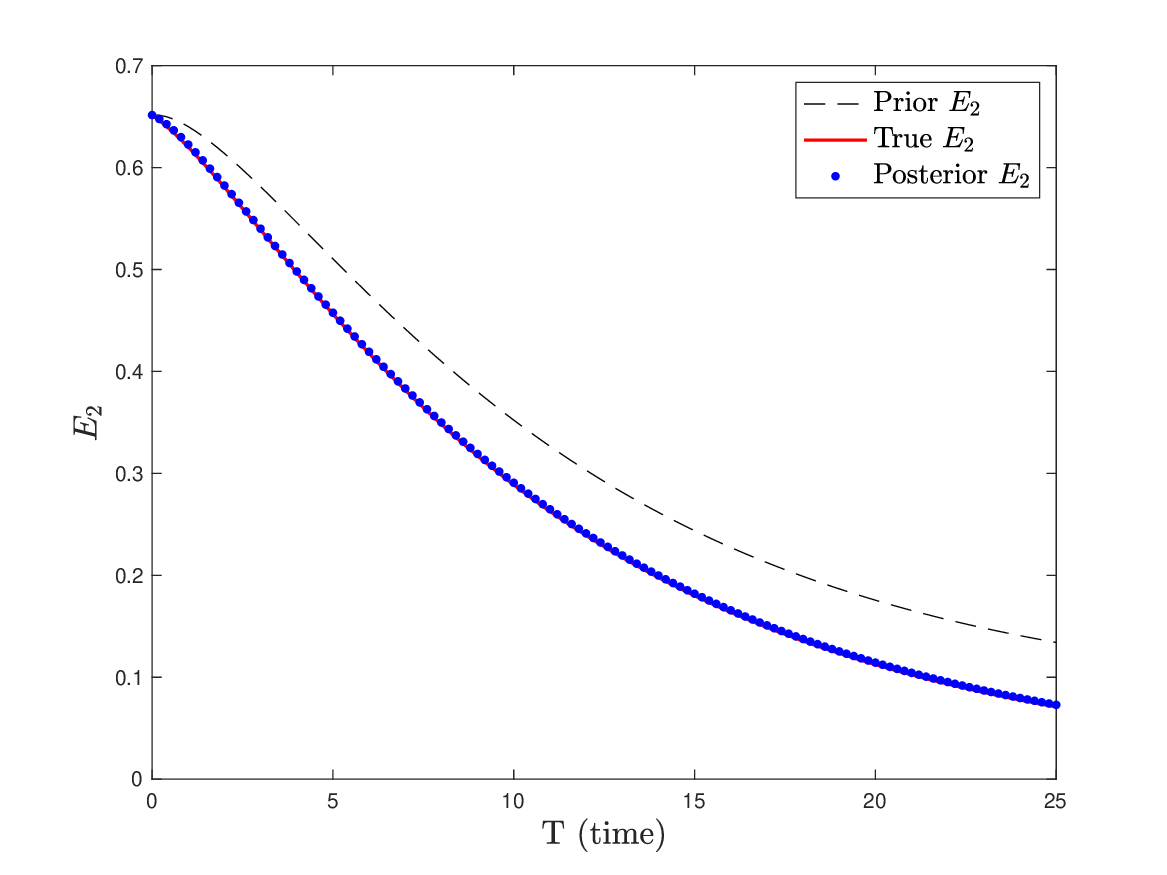}}
\hfil
{\includegraphics[width=0.42\textwidth]{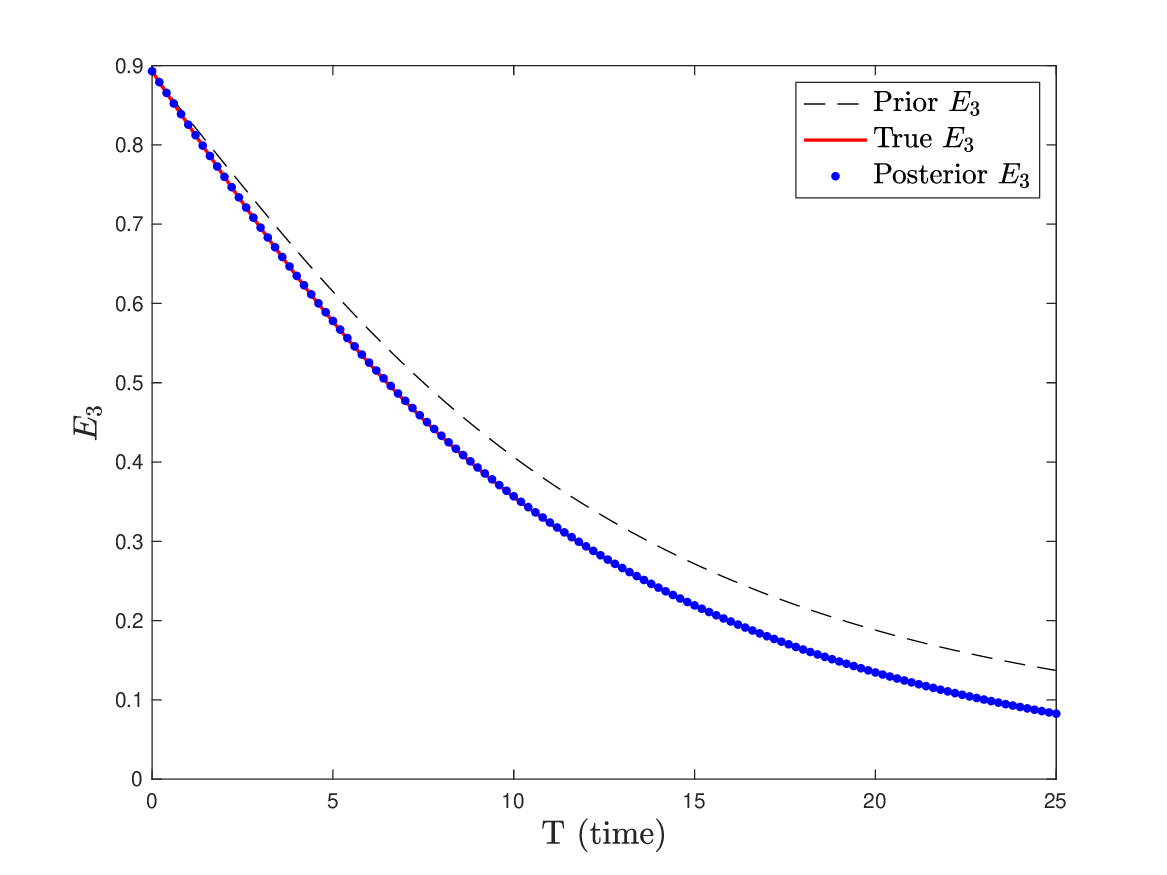}}

\medskip
{\includegraphics[width=0.42\textwidth]{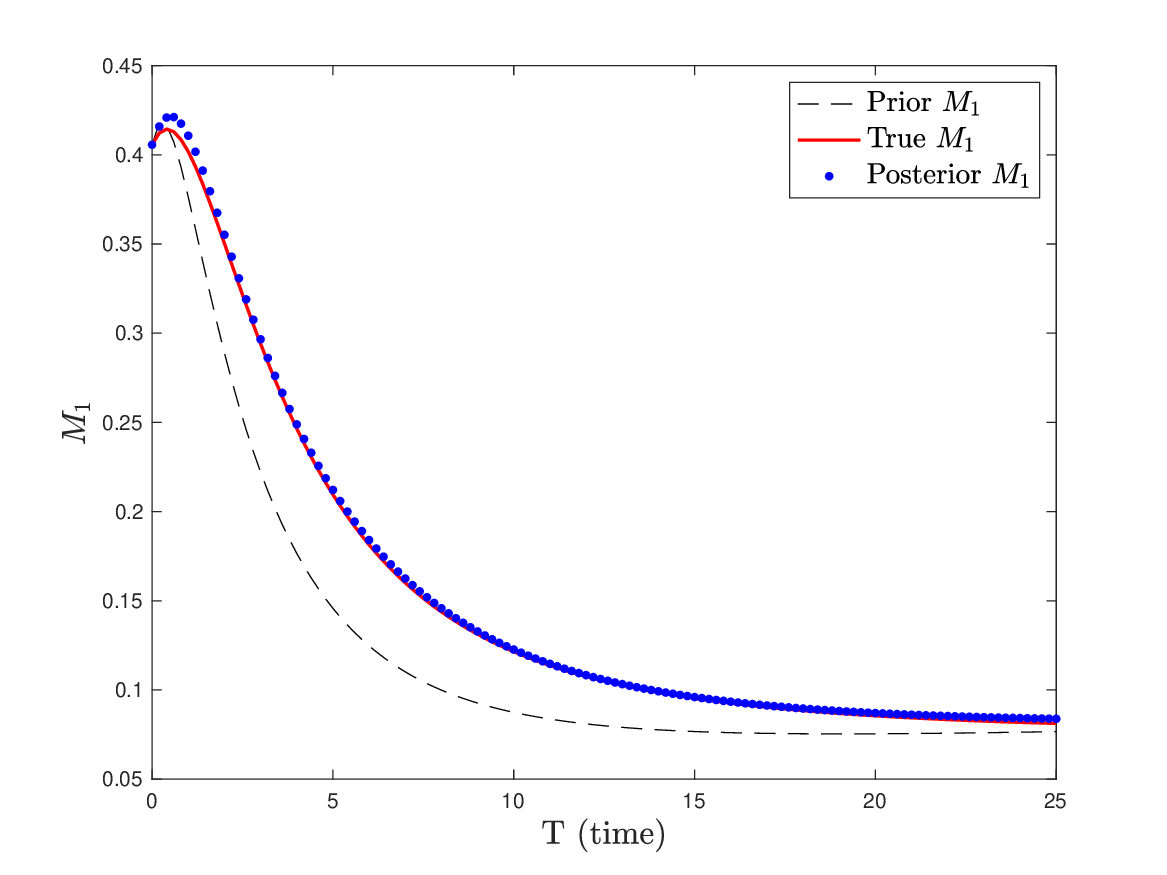}}
\hfil
{\includegraphics[width=0.42\textwidth]{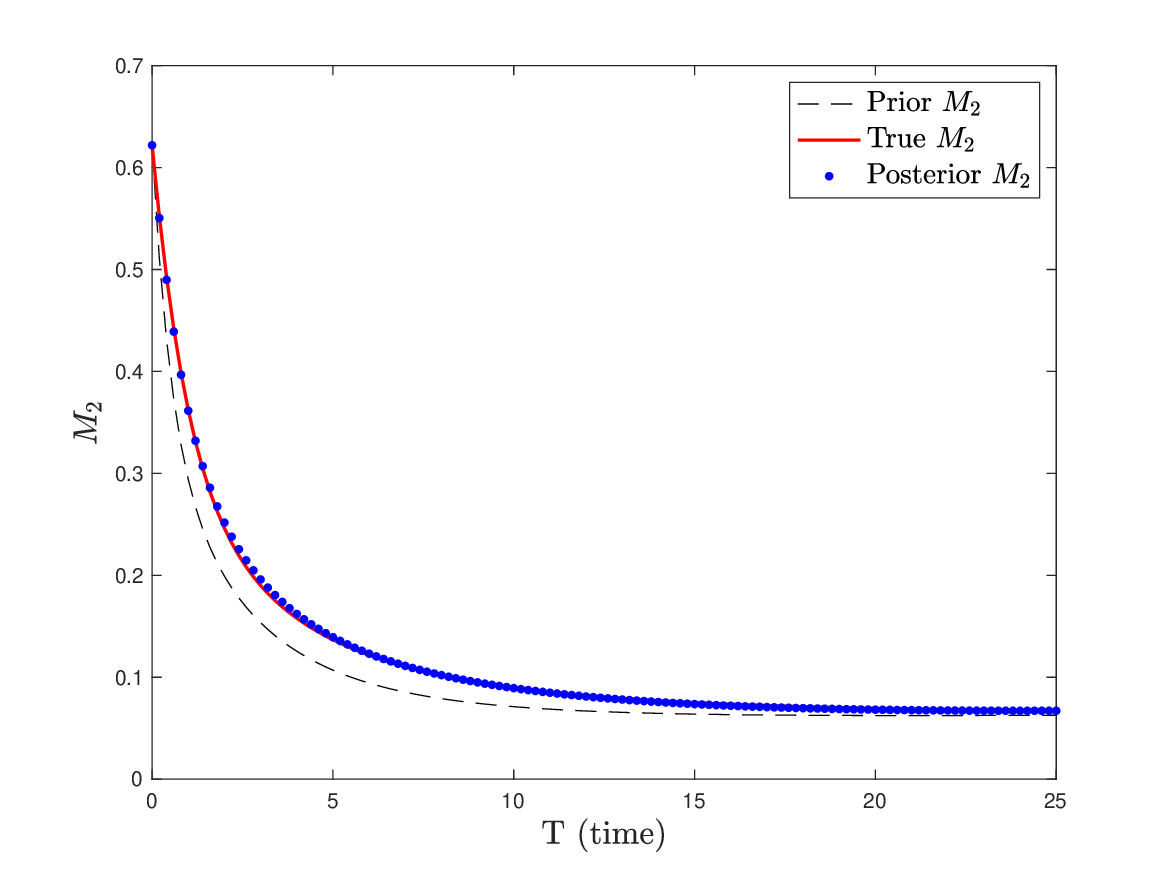}}
\caption{Metabolic pathway: system prediction up to $T=25$. Randomly selected initial condition: $G_1 = 0.253, G_2 = 0.679 , G_3 = 0.235, E_1 = 0.119, E_2 = 0.652, E_3 = 0.893, M_1 = 0.406, M_2 = 0.622$.}
    \label{metabolic_coarse_example}
    \end{figure}

\begin{figure}
\centering
 \includegraphics[width = 0.9\textwidth]{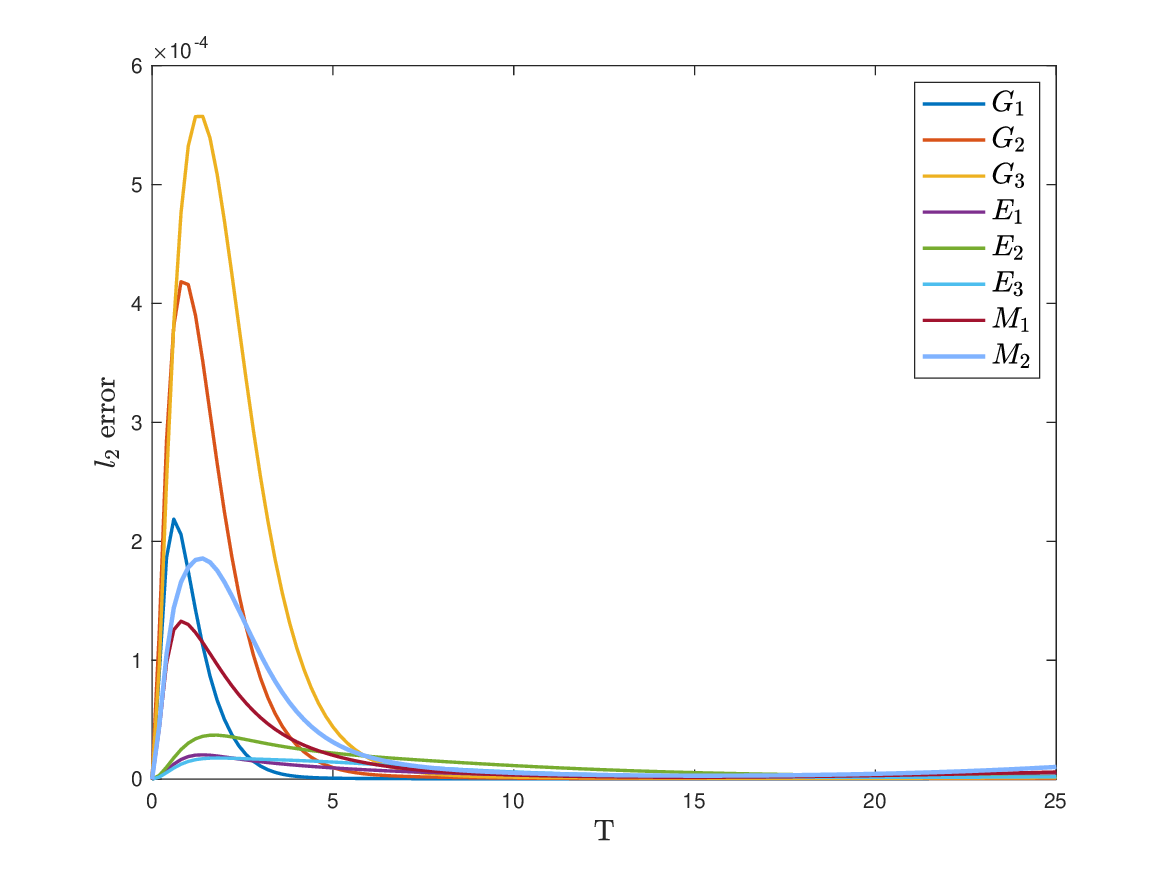}
    \caption{Metabolic pathway: average $l_2$ error over 100 test trajectories.}
    \label{metabolic_coarse_error}
\end{figure}

 In Figure \ref{metabolic_coarse_example}, predictions are presented up to $T = 25$ for a randomly selected initial condition. The average error for each of the $8$ state variables over $100$ test trajectories is depicted in Figure \ref{metabolic_coarse_error}. Given the challenging setting, the posterior DNN still provides clear improvement over the imperfect prior model and ability to provide accurate predictions.

%% file: Conclusion.tex
\section{Conclusion} \label{sec:conclusions}
We presented a model correcting framework for settings in which only scarce high-fidelity data may be available. Given an imperfect prior model and a small high-fidelity data set, we are able to obtain a posterior model capable of accurate predictions of the true system's state variables. 
This is obtained by correcting the prior DNN model via the deep learning technique of transfer learning. Our method contains two main novel features: (1) the form of the model correction is not assumed as additive or multiplicative but rather is conducted internally to the prior DNN model via transfer learning; (2) only a small (potentially coarse in time-resolution) high-fidelity data set is required for obtaining the posterior model, circumventing the computational challenge of DNN based learning methods in settings where only scarce high-fidelity data are available. We presented numerous numerical examples demonstrating the merit of the method. For future works, we plan to naturally extend this model correction method to partial differential equations as well as partially observed dynamical systems.


%% file: main.bbl
\begin{thebibliography}{10}

\bibitem{data_scarcity_article}
{\sc L.~Alzubaidi, J.~Bai, A.~Al-Sabaawi, J.~Santamaria, A.~Albahri, B.~Al-dabbagh, M.~Fadhel, M.~Manoufali, J.~Zhang, A.~Al-Timemy, Y.~Duan, A.~Abdullah, L.~Farhan, Y.~Lu, A.~Gupta, F.~Albu, A.~Abbosh, and Y.~Gu}, {\em A survey on deep learning tools dealing with data scarcity: definitions, challenges, solutions, tips, and applications}, Journal of Big Data, 10 (2023).

\bibitem{Validation_computer_models}
{\sc M.~Bayarri, J.~Berger, R.~Paulo, J.~Sacks, J.~Cafeo, C.~Cavendish, and J.~Tu}, {\em A framework for validation of computer models}, Technometrics, 49 (2007), pp.~138--154.

\bibitem{PriorConstraints_KennedyOHagan}
{\sc J.~Brynjarsdóttir and A.~O'Hagan}, {\em Learning about physical parameters, the importance of model discrepancy}, Inverse Problems, 30 (2014).

\bibitem{gResNet}
{\sc Z.~Chen and D.~Xiu}, {\em On generalized residual network for deep learning of unknown dynamical systems}, Journal of Computational Physics, 438 (2021).

\bibitem{inner_recurrence}
{\sc V.~Churchill and D.~Xiu}, {\em Learning fine scale dynamics from coarse observations via inner recurrence}, Journal of Machine Learning for Modeling and Computing,  (2022), pp.~61--77.

\bibitem{MLEforGP}
{\sc C.~Currin, T.~Mitchell, M.~Morris, and D.~Ylvisaker}, {\em Bayesian prediction of deterministic functions, with applications to the design and analysis of computer experiments}, Journal of the American Statistical Association, 86 (1991), pp.~953--963.

\bibitem{PINN_symbolic_model_discrepancy}
{\sc B.~Eastman, L.~Podina, and M.~Kohandel}, {\em A pinn approach to symbolic differential operator discovery with sparse data}, In: The Symbiosis of Deep Learning and Differential Equations II,  (2022).

\bibitem{Kutz_discrepancy_model}
{\sc M.~Ebers, K.~Steele, and J.~Kutz}, {\em Discrepancy modeling framework: learning missing physics, modeling systemic residuals, and disambiguating between deterministic and random effects}, SIAM Journal on Applied Dynamical Systems, 23 (2024).

\bibitem{Multi-fid_stochastic_collocation}
{\sc M.~Eldred, L.~Ng, M.~Barone, and S.~Domino}, {\em Multifidelity uncertainty quantification using spectral stochastic discrepancy models}, In: Handbook of Uncertainty Quantification, R.Ghanem D.Higdon and H. Owhadi,  (2017), pp.~1--45.

\bibitem{engl2009inverse}
{\sc e.~a. Engl, Heinz~W}, {\em Inverse problems in systems biology}, Inverse Problems, 25 (2009).

\bibitem{Imagetrans}
{\sc Y.~Gao and K.~Mosalam}, {\em Deep transfer learning for image-based structural damage recognition}, Computer-Aided Civil and Infrastructure Engineering, 33 (2018).

\bibitem{UQHandbook}
{\sc R.~Ghanem, D.~Higdon, H.~Owhadi, and et~al}, {\em Handbook of Uncertainty Quantification}, Springer Cham, 2017.

\bibitem{GP_params_noninform_priors}
{\sc R.~Haylock and A.~O'Hagan}, {\em On inference for outputs of computationally expensive algorithms with uncertainty on the inputs}, Bayesian Statistics, 5 (1996), pp.~629--637.

\bibitem{Xiu2016ModelCorrection}
{\sc Y.~He and D.~Xiu}, {\em Numerical strategy for model correction using physical constraints}, Journal of Computational Physics, 313 (2016), pp.~617--634.

\bibitem{Bayesian_MCMC_high_dim_output}
{\sc D.~Higdon, J.~Gattiker, B.~Williams, and M.~Rightley}, {\em Computer model calibration using high-dimensional output}, Journal of the American Statistical Association, 103 (2008), pp.~570--583.

\bibitem{Combining_field_data_and_computer_sims}
{\sc D.~Higdon, M.~Kennedy, J.~Cavendish, J.~Cafeo, and R.~Ryne}, {\em Combining field data and computer simulations for calibration and prediction}, SIAM Journal of Scientific Computation, 26 (2004), pp.~448--466.

\bibitem{Stats_engineering_models}
{\sc V.~Joseph and S.~Melkote}, {\em Statistical adjustments to engineering models}, Journal of Quality Tech., 41 (2009), pp.~362--375.

\bibitem{Statistical_calibration_embed_2019}
{\sc S.~K., X.~Huan, and N.~Habib}, {\em Embedded model error representation for bayesian model calibration}, International Journal of Uncertainty Quantification, 9 (2019), pp.~365--394.

\bibitem{Statistical_calibration_embed}
{\sc S.~K., H.~Najm, and R.~Ghanem}, {\em On the statistical calibration of physical models}, International Journal of Chemical Kinetics, 47 (2015).

\bibitem{Bayesian_MCMC_ANOVA}
{\sc C.~G. Kaufman and S.~R. Sain}, {\em Bayesian function {ANOVA} modeling using gaussian process prior distributions}, Bayesian Analysis, 5 (2010), pp.~123--149.

\bibitem{KennedyOHagan}
{\sc M.~C. Kennedy and A.~O'Hagan}, {\em Bayesian calibration of computer models}, Journal of the Royal Statistical Society: Series B (Statistical Methodology), 63 (2001), pp.~425--464.

\bibitem{transfer_learning_neyshabur}
{\sc B.~Neyshabur, H.~Sedghi, and C.~Zhang}, {\em What is being transferred in transfer learning?}, Advances in Neural Information Processing Systems, 33 (2020), pp.~512--523.

\bibitem{GP_params_conjugate_priors}
{\sc J.~Oakley and A.~O'Hagan}, {\em Probablistic sensitivity analysis of complex models; a bayesian approach}, Journal of the Royal Statistical Society: Series B (Statistical Methodology), 66 (2004).

\bibitem{Hierarchical_modeling_Qian}
{\sc Z.~Qian and C.~Wu}, {\em Bayesian hierarchical modeling for integration low-accuracy and high-accuracy experiments}, Technometrics, 50 (2008), pp.~192--204.

\bibitem{resnet}
{\sc T.~Qin, K.~Wu, and D.~Xiu}, {\em Data driven governing equations approximation using deep neural networks}, Journal of Computational Physics, 395 (2019).

\bibitem{transfer_learning_raghu}
{\sc M.~Raghu, C.~Zhang, J.~Kleinberg, and S.~Bengio}, {\em Transfusion: Understanding transfer learning for medical imaging}, Advances in Neural Information Processing Systems,  (2019), pp.~3342--3352.

\bibitem{NLPtrans}
{\sc S.~Ruder, M.~Peters, and S.~Swayamdipta}, {\em Transfer learning in natural language processing}, Proceedings of the 2019 Conference of the North American Chpater of the Association for Computational Linguistics,  (2019).

\bibitem{DeepOLTransNet}
{\sc G.~Somdatta, K.~Kontolati, M.~D. Shields, and G.~E. Karniadakis}, {\em Deep transfer operator learning for partial differential equations under conditional shift}, Nature Machine Intelligence, 4 (2022).

\bibitem{Bayesian_validation_Wang}
{\sc S.~Wang, W.~Chen, and K.~Tsui}, {\em Bayesian validation of computer models}, Technometrics, 51 (2009), pp.~439--451.

\bibitem{Bayesian_MCMC_chaotic_time_series}
{\sc D.~Williamson and A.~T. Blaker}, {\em Evolving bayesian emulators for structured chatoic time series, with application to large climate models}, SIAM/ASA Journal of Uncertainty Quantification, 2 (2014), pp.~1--28.

\bibitem{Structural_error_complex_dynamical_systems}
{\sc J.-L. Wu, M.~E. Levine, T.~Schneider, and A.~Stuart}, {\em Learning about structural errors in models of complex dynamical systems}, Journal of Computational Physics, 513 (2024).

\bibitem{transfer_learning_yosinski}
{\sc J.~Yosinski, J.~Clune, Y.~Bengio, and H.~Lipson}, {\em How transferable are features in deep neural networks?}, Advances in Neural Information Processing Systems,  (2014), pp.~3320--3328.

\bibitem{Transnet}
{\sc Z.~Zhang, F.~Bao, L.~Ju, and G.~Zhang}, {\em Transnet: Transferable neural networks for partial differential equations}, Journal of Scientific Computing, 99 (2024).

\bibitem{Alzheimers_application_model_discrepancy}
{\sc Z.~Zhang, Z.~Zou, E.~Kuhl, and G.~Karniadakis}, {\em Discovering a reaction-diffusion model for alzheimer's disease by combining pinns with symbolic regression}, Computer Methods in Applied Mechanics Engineering, 419 (2024).

\bibitem{PINNs_misspecification}
{\sc Z.~Zou, X.~Meng, and G.~E. Karniadakis}, {\em Correcting model misspecfification in physics-informed neural networks (pinns)}, Journal of Computational Physics, 505 (2024).

\end{thebibliography}


\clearpage
